\documentclass[letterpaper]{article} 
\usepackage[preprint]{aaai2027}  
\usepackage[hyphens]{url}  
\usepackage{graphicx} 
\usepackage{natbib}  
\usepackage{caption} 
\usepackage{booktabs}
\usepackage{multirow}
\usepackage{amssymb} 

\usepackage{xspace}
\DeclareRobustCommand{\ourmethod}{\textsc{PosterVisor}\xspace}
\newcommand{\PosterVisorAuthors}{%
    Runze Li\textsuperscript{\rm 1,2},
    Yukun Zhao\textsuperscript{\rm 3},
    Can Xu\textsuperscript{\rm 4},
    Yucheng Shen\textsuperscript{\rm 5},\\
    Shuaiqiang Wang\textsuperscript{\rm 1},
    Jianmin Wu\textsuperscript{\rm 1},
    Lingyong Yan\textsuperscript{\rm 1}\corresponding,
    Dawei Yin\textsuperscript{\rm 1}%
}

\newcommand{\PosterVisorAffiliations}{%
    \textsuperscript{\rm 1}Baidu Inc., Beijing, China\\
    \textsuperscript{\rm 2}Harbin University of Science and Technology, Harbin, China\\
    \textsuperscript{\rm 3}Shandong University, Jinan, China\\
    \textsuperscript{\rm 4}East China Normal University, Shanghai, China\\
    \textsuperscript{\rm 5}Soochow University, Suzhou, China%
}

\title{From Transient Prompts to Persistent Control: Scientific Poster \\ Generation via Recursive Semantic--Geometric Contracts}
\author{\PosterVisorAuthors}
\affiliations{\PosterVisorAffiliations}

\begin{document}

\maketitle

\begin{abstract}
Scientific poster generation distills a multimodal paper into a single-page visual artifact, forcing strict trade-offs between informational coverage and readability under a fixed spatial budget. Existing methods pass plans as transient prompts and validate individual stages in isolation. This strategy causes requirements to drift across content and layout modules, and previous checks to be silently invalidated.
We introduce \ourmethod, a control framework that shifts poster generation from transient prompts to persistent control. An Orchestrator grounds rubrics in the paper and visual assets, compiling them into a Semantic--Geometric Contract (SGC) that binds claims and sources to required visuals, budgets, and spatial commitments. Only fully instantiated records become executable assertions; other usable requirements remain soft guidance. Recursive Contract Enforcement (RCE) dynamically triggers checks across stages as evidence emerges. Crucially, during repairs, RCE rechecks affected checkpoint states, preventing repair-induced regressions from propagating silently. We instantiate \ourmethod in HTML/CSS and editable PPTX generators. On the 100-paper Paper2Poster benchmark, \ourmethod-PPT improves observed mean poster-grounded QA accuracy over PosterGen (64.47\% vs.\ 58.53\%) and is preferred by human judges in 72.5\% of non-tied pairwise comparisons (95\% CI, 61.6--83.4\%). A secondary 30-paper study also yields higher VLM Overall and PaperQuiz means. These results support rubric-compiled contracts and stage-conditioned enforcement for controllable poster synthesis.

\end{abstract}


\section{Introduction}

Scientific posters provide a compact visual medium for presenting research papers, distilling a full paper into a single-page narrative~\cite{qiang2017poster,zhong2025sciposter,tanaka2024scipostlayout}. 
Automating this process goes beyond conventional text summarization: under a fixed spatial budget, a system must jointly select scientific claims, retain supporting evidence, and coordinate text, visuals, and layout to clearly convey the paper's core message while keeping its visual evidence legible~\cite{sun2025p2p,pang2025paper2poster,zhang2025postergen,choi2025posterforest}.

Recent LLM- and VLM-based systems frame scientific poster generation as a multi-stage process. They combine content–layout planning, specialized agents, hierarchical or editable representations, and visual feedback to coordinate content, layout, and revision (Pang et al. 2025; Sun et al. 2026; Zhang et al. 2026; Choi et al. 2026). More recent work improves efficiency and auditability through token compression, targeted violation detection, provenance tracking, and geometry checks (Tang et al. 2026; Yang et al. 2026). However, these pipelines still bind plans and checks to individual stages rather than maintaining a shared, paper-specific acceptance state throughout generation.



This limitation creates two control challenges. First, semantic and spatial requirements are frequently weakened as a poster evolves from a textual plan to structured content and rendered outputs. Early plans merely specify what to generate, failing to bind claims, evidence, and spatial allocations into a persistent specification. Second, since requirements become observable at different stages, later modifications can silently invalidate earlier checks. Without tracking the scope and dependencies of each repair, the system cannot determine which results remain valid.

To address these challenges, we introduce \ourmethod, an Orchestrator-centered framework that turns poster generation from transient prompts to persistent contracts. It turns paper-specific planning decisions into a \textit{Semantic--Geometric Contract} (SGC) shared across the content generation, layout construction, and rendering stages. Starting from a fixed catalog of poster-quality criteria, the Orchestrator grounds relevant criteria in the source paper and its visual assets, binding narrative claims and source evidence to required visuals, content budgets, and coarse spatial commitments. Requirements with explicit targets and available validators are compiled into executable assertions, while other useful requirements are retained as soft guidance. Because these requirements become observable at different stages, \textit{Recursive Contract Enforcement} (RCE) evaluates each executable assertion when the evidence needed to check them becomes available. When an assertion fails, RCE routes the violation to a scoped and bounded repair and then revalidates the assertions that may have been affected by the change. SGC therefore limits the drift of semantic and spatial commitments, while RCE prevents affected validation results from being silently treated as valid after repair. Together, SGC and RCE extend stage-local planning and checking into persistent, repair-aware control throughout poster generation.

We instantiate \ourmethod in two systems: a P2P-derived HTML/CSS generator and a PosterGen-derived editable-PPTX generator. On the primary benchmark of 100 papers, both implementations obtain higher observed VLM Overall and Raw PaperQuiz means than their matched baselines and tie for the highest displayed automatic VLM Overall of 3.82. \ourmethod-PPT improves Raw PaperQuiz by 5.94 points and is preferred to PosterGen in 72.5\% of non-tied human comparisons (95\% CI, 61.6--83.4\%). Across four additional model backbones, \ourmethod-HTML improves both objectives in all four settings, while \ourmethod-PPT improves each objective in three of the four settings. Archived execution logs further show that all 159 recorded deterministic repair events are followed by rule-based rechecking, providing direct evidence of the intended enforcement behavior. A secondary study on 30 papers provides additional evidence of transferability.

Our contributions are summarized as follows:
\begin{itemize}
    \item We formulate scientific poster generation as a cross-representation control problem in which coupled semantic, evidential, and spatial commitments become observable at different stages and may be invalidated by subsequent repairs.
    \item We introduce a rubric-grounded \textit{Semantic--Geometric Contract} that preserves paper-specific commitments as a shared acceptance state, together with \textit{Recursive Contract Enforcement}, which evaluates executable assertions at appropriate checkpoints and revalidates affected assertions after scoped and bounded repairs.
    \item We instantiate the framework as HTML/CSS and editable PPTX generators, and demonstrate its effectiveness across two benchmarks in terms of output quality, cross-model robustness, human preference, and recorded control behavior.
\end{itemize}

\section{Related Work}

\paragraph{Scientific Poster Generation.}
Early scientific-poster systems select and arrange content using learned statistics or neural components~\cite{qiang2017poster,xu2022posterbot}. Subsequent resources support poster generation, layout analysis, structural parsing, and summarization~\cite{zhong2025sciposter,tanaka2024scipostlayout,tanaka2026scipostlayouttree,saxena2025postersum}. Recent LLM/VLM pipelines use structured intermediate representations to transform parsed document content into editable visual structures for posters, slides, and webpages~\cite{pang2025paper2poster,sun2025p2p,zhang2025postergen,choi2025posterforest,sun2021d2s,zheng2025pptagent,ma2025autopage}. A typical agentflow decomposes the task into paper parsing, content and figure selection, narrative planning, content writing, layout construction, rendering, and output review~\cite{pang2025paper2poster,sun2025p2p,zhang2025postergen,choi2025posterforest,tang2026efficientpostergen}. PosterHarness further advances contract-oriented control through placeholder-first planning, staged QA, and deterministic source-figure composition~\cite{yang2026posterharness}. Any2Poster and ResearchStudio-Reel broaden the task to diverse inputs and editable multi-artifact outputs~\cite{vinaykumar2026any2poster,xiao2026researchstudio}.

\paragraph{Rubric-Guided Evaluation and Recursive Enforcement.}
Rubrics decompose open-ended quality into interpretable criteria for evaluation, diagnosis, and reward or verifier design; recent work also automates their synthesis and extraction~\cite{hashemi2024llmrubric,sharma2025researchrubrics,gunjal2025rubrics,he2025advancedif,shao2025drtulu,liu2026openrubrics,xie2026autorubric,li2026rubrichub}. Prior poster-generation systems likewise employ fine-grained criteria, stage-specific checkers, rendered-image critics, deterministic layout tests, and rejection rules~\cite{pang2025paper2poster,sun2025p2p,zhang2025postergen,choi2025posterforest,tang2026efficientpostergen,yang2026posterharness}. These mechanisms address local defects, but their criteria usually remain attached only to the current artifact at a particular stage. For example, a plan may require a key figure to be both included and readable. A content checker may confirm its inclusion, while an overflow checker may confirm that the layout fits; yet the figure may remain too small to convey its evidence. Enlarging it may then cause text overflow unless the affected constraints are rechecked. These mechanisms therefore provide useful local feedback but offer limited cross-stage persistence and repair-aware revalidation.

\section{Problem Formulation}

Given a multimodal source paper $P$, a fixed poster-quality rubric catalog $R$, and a target output format $f$, we formulate controlled poster generation as \[
(P,R,f)\longmapsto(X,Y,L),
\]
where $X$ is an editable artifact, $Y$ is its rendered poster, and $L$ is an audit log. Requirements instantiated from $P$ and $R$ under format $f$ combine scientific content, source evidence, and spatial presentation. The control problem is to maintain and evaluate this paper-specific requirement set across generated content, editable geometry, and rendered output, recognizing that requirements become verifiable at different stages and repairs may affect previously satisfied constraints.
\section{Method}

In this section, we first summarize \ourmethod's end-to-end workflow and then describe its two core components: Semantic--Geometric Contract (SGC) construction and Recursive Contract Enforcement (RCE).

\begin{figure*}[t]
\centering
\includegraphics[width=1.0 \textwidth]{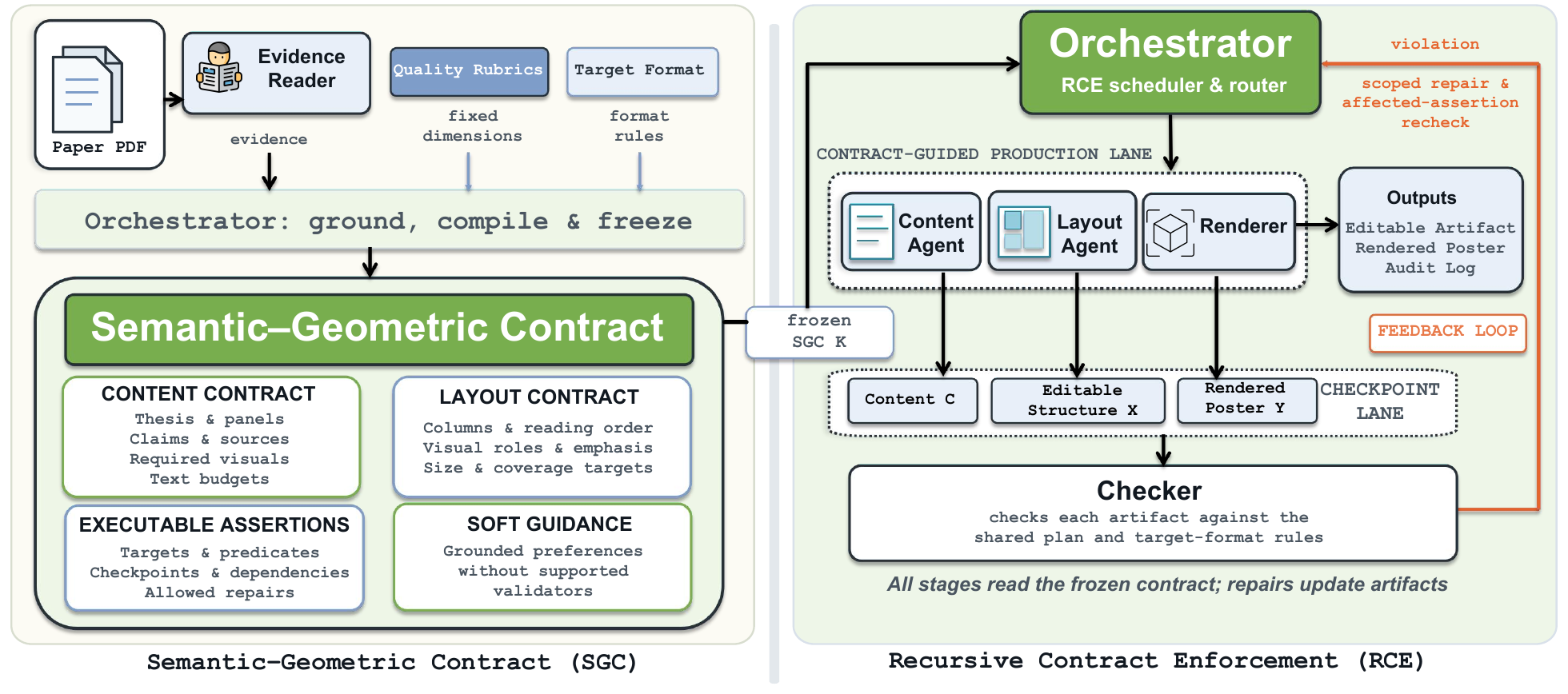}
\caption{Overview of \ourmethod. The Evidence Reader extracts traceable paper evidence, and the Orchestrator grounds rubric dimensions to construct a persistent SGC. Content and layout modules generate artifacts under the frozen contract. RCE evaluates assertions as state becomes available, routes violations to bounded repairs, and rechecks affected assertions.}
\label{fig:postervisor_overview}
\end{figure*}

\subsection{Overview}

\ourmethod first uses an Evidence Reader to transform the source paper $P$ into indexed text $T$ and a provenance-preserving visual inventory $A$. The Orchestrator uses $(T,A)$ to determine the thesis, priorities, and spatial allocation. It grounds applicable criteria from $R$ to freeze the SGC $K$, binding claims to required visuals, budgets, and spatial commitments. Guided by $K$, the Content Agent writes the poster content $C$. The Layout Agent combines the checked content, selected visuals, and spatial requirements into the editable artifact $X$, which a format-specific renderer converts into $Y$. At the content, editable-layout, and rendered-output checkpoints, the Checker evaluates applicable assertions against the corresponding artifact state and returns structured violations. Under RCE, the Orchestrator routes each actionable violation to an allowed repair operator, and the Checker rechecks the assertions affected by the repair. We instantiate this shared SGC/RCE abstraction in a P2P-derived HTML/CSS generator and a PosterGen-derived editable-PPTX generator. Figure~\ref{fig:postervisor_overview} summarizes the workflow.

\subsection{Semantic--Geometric Contract Construction}

The fixed catalog $R$ defines the general quality dimensions considered in our experiments: scientific-content coverage, visual-evidence sufficiency, information density, visual readability, figure--text balance, and spatial balance. For a source paper $P$, the Orchestrator grounds the applicable dimensions in the indexed text $T$ and visual inventory $A$ and represents the resulting paper-specific requirements as an SGC,
$K=(K_c,K_l,\mathcal{E},\mathcal{S})$.
$K_c$ stores the poster thesis, ordered panels, claims or content intents, source anchors, required visual evidence, and text budgets. $K_l$ stores panel-to-column assignments, reading order, spatial emphasis, visual roles, key-visual protection, and paper-specific size or coverage targets. $\mathcal{E}$ contains fully instantiated, checkable assertions, whereas $\mathcal{S}$ retains grounded records that can guide generation but are not executable. Each panel record thereby binds scientific intent and evidence to a content budget and a coarse spatial commitment. Only assertions in $\mathcal{E}$ determine verified contract compliance. Format-wide measurement rules, default tolerances, checkpoints, and admissible repair operators are specified separately by the format policy $\Theta_f$.

Planning first produces a candidate record set $\mathcal{G}$. For each applicable rubric dimension, grounding yields zero or more records
$g_j=\langle r_j,q_j,\widehat{\eta}_j,p_j\rangle$,
where $r_j$ identifies the criterion, $q_j$ is a typed target query, $\widehat{\eta}_j$ is a required value or policy key, and $p_j\subseteq T\cup A$ records paper provenance. The Orchestrator also adds schema-derived records for required sections, evidence bindings, and budgets. Normalization resolves $q_j$ and $\widehat{\eta}_j$ against the draft contract and $\Theta_f$, producing either
$g'_j=\langle r_j,\tau_j,\eta_j,p_j\rangle$
or an explicitly unresolved record in $\mathcal{G}'$.

Each executable assertion has the form
\[
e_i=\langle c_i,\tau_i,\eta_i,\phi_i,s_i,D_i,\mathcal{U}_i\rangle ,
\]
where $c_i$ is a violation code, $\tau_i$ is a typed target, $\eta_i$ is the expected value or bound, $\phi_i$ is the checking predicate, and $s_i$ is its checkpoint. $D_i$ lists read dependencies, and $\mathcal{U}_i$ is a repair allow-list that may be empty. The format policy supplies a finite catalog of typed predicate templates. A policy key is not executable by itself: before a record enters $\mathcal{E}$, its target, expectation, and checking rule must be materialized by the run's contract state and format-policy configuration. Accordingly, for $g'_j\in\mathcal{G}'$, the compiler emits
$e_i=C_f(g'_j;K_c,K_l,\Theta_f)\in\mathcal{E}$
only when $\tau_i$ and $\eta_i$ are resolved and a compatible template defines
$(\phi_i,s_i,D_i,\mathcal{U}_i)$. Otherwise, the record enters $\mathcal{S}$ only if it remains usable as generation guidance; unusable records are retained as unresolved in $L$. 

To illustrate this compilation mechanism, consider a rubric requiring ``visual balance.'' Grounding might resolve this abstract concept to extracting Figure 3 from the paper. During compilation, if the format policy $\Theta_f$ supports measuring image area, this becomes an executable assertion in $\mathcal{E}$ (e.g., ``Figure 3 must occupy $\ge 15\%$ of the column''). Conversely, if the system cannot reliably validate a specific aesthetic style, that requirement degrades to soft guidance in $\mathcal{S}$, prompting the generator without forcing a hard check. For a compiled assertion,
$\phi_i(Z[D_i],\tau_i;\eta_i,\Theta_f)$ returns
$\{\textsc{pass},\textsc{fail},\bot\}$, where $\bot$ denotes an unavailable observation or checker failure and is never treated as a pass.


Indexed text, captions, visual references, asset geometry, and readability or importance estimates provide the evidence used to resolve record targets and expectations. The construction process is summarized as
\[
\begin{array}{rcl}
(T,A) & = & E(P),\\
(\widetilde K,\mathcal{G}) & = & \Pi(T,A;R),\\
(K_c,K_l,\mathcal{G}') & = & N(\widetilde K,\mathcal{G};T,A,\Theta_f),\\
(\mathcal{E},\mathcal{S}) & = & C_f(\mathcal{G}';K_c,K_l,\Theta_f).
\end{array}
\]
Here, $E$, $\Pi$, $N$, and $C_f$ denote evidence extraction, paper-specific planning and grounding, normalization, and assertion compilation, respectively; $\widetilde K$ is the draft contract produced during planning. Normalization resolves identifier, anchor, budget, and estimated-load conflicts before the final contract
$K=(K_c,K_l,\mathcal{E},\mathcal{S})$
is frozen. Any conflict that cannot be resolved remains a plan violation. After freezing, $K$ remains unchanged, and state-specific projections of the same contract are supplied to generation, checking, and repair. Format-specific modules may instantiate supplementary artifact-specific checks from $\Theta_f$ for diagnosis or auxiliary repair, but these checks neither modify $K$ nor expand $\mathcal{E}$.

\subsection{Recursive Contract Enforcement}
Recursive Contract Enforcement (RCE) evaluates executable assertions when their
required state is available at checkpoint $s_t$. Content checkpoints expose
sections, claims, evidence references, sources, and budgets. Layout checkpoints
expose structural feasibility and realized geometry. Within $Q_{s_t}$, executable assertions are evaluated either by deterministic validators for computable structural and geometric requirements or by task-scoped model predicates for properties that are difficult to fully formalize, such as plan coverage, figure--text consistency, and rendered content quality. Model outputs affect $Q_{s_t}$ only when they implement the predicate of an assertion already compiled into $\mathcal{E}$; other critic findings remain diagnostic or may trigger auxiliary repair without affecting contract compliance. For an artifact state $Z_t\in\{C,X,Y\}$, the Checker and repair router compute
\[
\begin{array}{rcl}
\mathcal{V}_t & = & Q_{s_t}(Z_t;\mathcal{E}_{s_t}(K),\Theta_f),\\
u_t & = & H(v_t,Z_t;K,\Theta_f)\in\mathcal{U}(v_t),\\
Z_{t+1} & = & \rho(u_t,Z_t;K,\Theta_f),\\
W_t & = & W_{\mathrm{decl}}(u_t)\cup
                 \mathrm{diff}(Z_t,Z_{t+1}),\\
\mathcal{I}_t & = &
 \{e_i\in\mathcal{E}:D_i\cap W_t\neq\varnothing\},\\
\Gamma_t & = & \mathrm{cl}_{K,\Theta_f}
 \bigl(\{e(v_t)\}\cup\mathcal{I}_t\bigr),
\end{array}
\]
for an actionable violation $v_t\in\mathcal{V}_t$. Here, $Q_{s_t}$ returns
violations among the assertions active at checkpoint $s_t$, and
$\mathcal{U}(v_t)$ is the failed assertion's nonempty repair allow-list. $W_t$
combines the repair's declared scope with the observed artifact difference.
$\mathcal{I}_t$ identifies directly affected assertions, while $\Gamma_t$ adds
the failed assertion $e(v_t)$ and closes this set under contract- and
format-specific dependencies. If a reliable structural difference is
unavailable, the Checker reruns all executable assertions over the modified
artifact state.

The closure $\Gamma_t$ additionally includes transitive geometry dependencies and global structural guards. When reliable field-level dependencies are unavailable, the full-checkpoint fallback conservatively includes all applicable assertions at the modified state. This dependency-aware rechecking is our key departure from single-pass self-correction, which may silently retain cascading errors. For example, after an image is shrunk to fix spatial overflow, the recheck scope includes image-text readability and adjacent alignment, allowing any resulting regression to be detected rather than silently accepted. After repair, the Checker reevaluates $\Gamma_t$, and a target is considered resolved only when its predicate passes.
If an assertion changes from \textsc{pass} on $Z_t$ to \textsc{fail} or $\bot$
on $Z_{t+1}$, RCE records a typed \textsc{regression} violation. The frozen
contract $K$ provides the comparison reference. Thus, the observed-delta
closure makes regressions detectable, whereas contract immutability alone
neither prevents nor repairs them.

Repair types depend on the violation, not the checkpoint (e.g., layout overflow may trigger geometric tweaks or content shortening). RCE uses deterministic operators for directly measurable failures and scoped model calls when regeneration is required. It continues only while the relevant verification results improve and stops when all executable assertions applicable to the final artifact return \textsc{pass}, no permitted repair remains, verification no longer improves, or the format-specific repair budget is exhausted. Contract compliance is assigned only when all applicable assertions pass. Otherwise, the system returns the latest available artifact and records failed or unresolved assertions, exhausted repairs, and detected regressions in $L$. Regression detection is limited by the declared dependencies, observed differences, and underlying checkers; soft guidance and unmodeled properties remain outside the compliance claim.

\section{Experiments}

\begin{table*}[!t]
\small
\centering

\scriptsize
\setlength{\tabcolsep}{1.45pt}
\resizebox{\textwidth}{!}{%
\begin{tabular}{@{}l*{15}{c}@{}}
\toprule
\multirow{3}{*}{Method}
& \multicolumn{9}{c}{VLM-as-Judge ($\uparrow$)}
& \multicolumn{6}{c}{PaperQuiz ($\uparrow$)} \\
\cmidrule(lr){2-10} \cmidrule(lr){11-16}
& \multicolumn{4}{c}{Aesthetic}
& \multicolumn{4}{c}{Information}
& \multirow{2}{*}{Overall}
& \multicolumn{3}{c}{Raw Accuracy}
& \multicolumn{3}{c}{Density-Augmented} \\
\cmidrule(lr){2-5} \cmidrule(lr){6-9}
\cmidrule(lr){11-13} \cmidrule(lr){14-16}
& Elem. & Layout & Engage. & Avg.
& Clarity & Content & Logic & Avg. & 
& Verb. & Interp. & Overall
& Verb. & Interp. & Overall \\
\midrule
Paper$^\dagger$
& 4.05 & 3.89 & 2.80 & 3.58 & 4.00 & 4.68 & 3.98 & 4.22 & 3.90
& 84.41 & 79.69 & 82.05 & 92.97 & 87.75 & 90.36 \\
GT Poster$^\dagger$
& 4.07 & 3.90 & 2.70 & 3.56 & 4.09 & 3.96 & 3.89 & 3.98 & 3.77
& 68.25 & 74.73 & 71.49 & 127.75 & 140.29 & 134.02 \\
\midrule
PosterAgent$^\ddagger$
& 3.92 & 3.05 & 2.98 & 3.32 & 3.98 & 3.97 & 3.55 & 3.83 & 3.58
& \underline{57.81} & 76.39 & \textbf{67.10} & \underline{111.40} & 147.29 & 129.34 \\
PosterGen$^\star$
& 3.59 & 2.98 & 2.81 & 3.13 & 4.14 & 3.93 & 3.32 & 3.80 & 3.46
& 43.90 & 73.15 & 58.53 & 87.71 & 146.19 & 116.95 \\
PPTAgent$^\star$
& 2.76 & 2.83 & 1.93 & 2.51 & 3.80 & 2.68 & 2.31 & 2.93 & 2.72
& 22.24 & 60.09 & 41.17 & 44.48 & 120.19 & 82.33 \\
P2P$^\star$
& \textbf{3.99} & 3.50 & 3.01 & 3.50 & 3.99 & 3.92 & 3.83 & 3.91 & 3.71
& 55.67 & 76.60 & 66.14 & 111.19 & 153.03 & \underline{132.11} \\
PosterForest$^\star$
& 3.96 & \underline{3.56} & 3.01 & \underline{3.51} & 3.90 & 3.91 & 3.70 & 3.84 & 3.67
& 51.05 & \textbf{80.42} & 65.73 & 101.99 & \textbf{160.74} & 131.37 \\
Any2Poster$^\star$
& 3.10 & 2.87 & 2.39 & 2.79 & 3.76 & 3.69 & 3.29 & 3.58 & 3.18
& 49.91 & \underline{79.89} & 64.90 & 99.59 & \underline{159.45} & 129.52 \\
EfficientPosterGen$^\star$
& 3.18 & 2.99 & 2.21 & 2.79 & 4.09 & 3.96 & 3.43 & 3.83 & 3.31
& 55.08 & 76.71 & 65.89 & 109.98 & 153.18 & 131.58 \\
PosterHarness$^\star$
& 3.42 & 3.40 & \underline{3.02} & 3.28 & 4.29 & \textbf{4.14} & \underline{4.14} & \underline{4.19} & 3.74
& 55.50 & 76.68 & 66.09 & 110.94 & 153.28 & \underline{132.11} \\
\midrule
\ourmethod-PPT
& \underline{3.97} & \textbf{3.76} & \textbf{3.09} & \textbf{3.61} & \underline{4.45} & \underline{4.05} & 3.59 & 4.03 & \textbf{3.82}
& 53.02 & 75.91 & 64.47 & 106.04 & 151.83 & 128.93 \\
\ourmethod-HTML
& 3.73 & 3.45 & 2.96 & 3.38 & \textbf{4.62} & 3.92 & \textbf{4.25} & \textbf{4.26} & \textbf{3.82}
& \textbf{58.23} & 75.40 & \underline{66.82} & \textbf{115.95} & 150.21 & \textbf{133.08} \\
\bottomrule
\end{tabular}
}
\caption{Results on the 100-paper Paper2Poster benchmark. $\dagger$ marks values reported by Paper2Poster, $\ddagger$ marks our LaTeX PosterAgent reproduction, and $\star$ marks our other reproductions. Paper and GT Poster are reference inputs excluded from automatic-method ranking. Bold and underline indicate the best and second-best automatic results; rounded ties share a mark.}
\label{tab:main_results}
\end{table*}

\subsection{Experimental Setup}

\paragraph{Datasets.}
Our primary benchmark is Paper2Poster~\cite{pang2025paper2poster}, containing 100 source papers, author posters, and factual QA data. We additionally use a 30-paper secondary evaluation set~\cite{zhang2025postergen}. Lacking PaperQuiz data for this set, we reproduce the Paper2Poster protocol to generate 50 verbatim and 50 interpretive questions per paper. All generators in the main comparison use GPT-4o, whereas the secondary comparison uses GPT-4.1.

\paragraph{Baselines.}
We compare against eight automatic systems: \textbf{PosterAgent}~\cite{pang2025paper2poster}, \textbf{P2P}~\cite{sun2025p2p}, \textbf{PosterGen}~\cite{zhang2025postergen}, \textbf{PPTAgent}~\cite{zheng2025pptagent}, \textbf{PosterForest}~\cite{choi2025posterforest}, \textbf{Any2Poster}~\cite{vinaykumar2026any2poster}, \textbf{EfficientPosterGen}~\cite{tang2026efficientpostergen}, and \textbf{PosterHarness}~\cite{yang2026posterharness}. P2P and PosterGen are the respective base generators for \ourmethod-HTML and \ourmethod-PPT, enabling direct within-generator comparisons. Superscripts in Table~\ref{tab:main_results} distinguish reported reference values from our reproductions.

\paragraph{Evaluation Metrics.}
Following Paper2Poster~\cite{pang2025paper2poster}, \textbf{VLM-as-Judge} uses GPT-4o to score six criteria from 1 to 5. Element quality, layout balance, and engagement form \textit{Aesthetic}; clarity, content completeness, and logical flow form \textit{Information}; \textit{Overall} averages all six. \textbf{PaperQuiz} measures source-information recoverability with 50 verbatim and 50 interpretive questions answered by GPT-4o, GPT-4o-mini, and o3. The Paper reference provides up to 20 rendered pages, whereas each poster method provides one image. Unsupported answers count as incorrect, and reader scores are averaged per sample. Density augmentation is computed before cross-paper averaging as $s_A=s_R(1+1/\max(1,l/w))$, where $l$ is the OCR-text token count and $w=1335.5$ is the median over re-evaluated author posters. These criteria are external output-quality measures rather than the generation-time rubric catalog $R$ or direct tests of contract compliance.

\paragraph{Implementation Details.}
Both realizations share the rubric catalog $R$ and the SGC/RCE abstraction but use format-specific artifact states, validators, and repair operators. \ourmethod-HTML checks generated content, DOM structure, and Playwright-rendered geometry. It uses a 650-word budget, balanced density, at least three required visuals, up to two content and DOM repair rounds, and three layout--render--check attempts. \ourmethod-PPT checks its panel blueprint, editable PPTX geometry, and LibreOffice-rendered output and permits at most three repair--render rounds. Only executable validators determine contract compliance; if a repair budget is exhausted, the latest artifact is returned with its remaining violations recorded rather than being marked compliant. Each evaluated system produces one poster per paper without candidate selection; within each benchmark, newly evaluated outputs share the same judge and PaperQuiz protocol. Each run retains the frozen SGC, structured violations, applied repair patches, geometry reports, unresolved assertions, and final artifacts for audit.

\subsection{Main Results}

Table~\ref{tab:main_results} shows higher observed means for both matched comparisons. \ourmethod-HTML raises VLM Overall from 3.707 to 3.822 and Raw PaperQuiz Overall from 66.14 to 66.82, while attaining the highest Density-Augmented Overall of 133.08. \ourmethod-PPT raises the corresponding VLM and Raw PaperQuiz scores from 3.460 to 3.818 and from 58.53 to 64.47. Both realizations round to the highest displayed automatic VLM Overall of 3.82. Fine-grained results show complementary strengths: \ourmethod-HTML leads clarity, logical flow, and aggregate Information, whereas \ourmethod-PPT leads engagement and improves most strongly over its base in Aesthetic.

Cross-model experiments with Qwen3.7-Plus, GPT-5.5, Gemini-3.1-Pro-Preview, and Claude-Opus-4.8 further test whether the control abstraction depends on GPT-4o. \ourmethod-HTML improves both VLM Overall and Raw PaperQuiz over P2P in all four settings, by 0.019--0.127 VLM points and 1.56--3.90 QA points. \ourmethod-PPT improves VLM Overall in three settings and Raw PaperQuiz in three settings, revealing stronger model-dependent content--presentation trade-offs. Full results appear in the supplement.

\begin{figure*}[!t]
\centering
\includegraphics[width=0.99\textwidth]{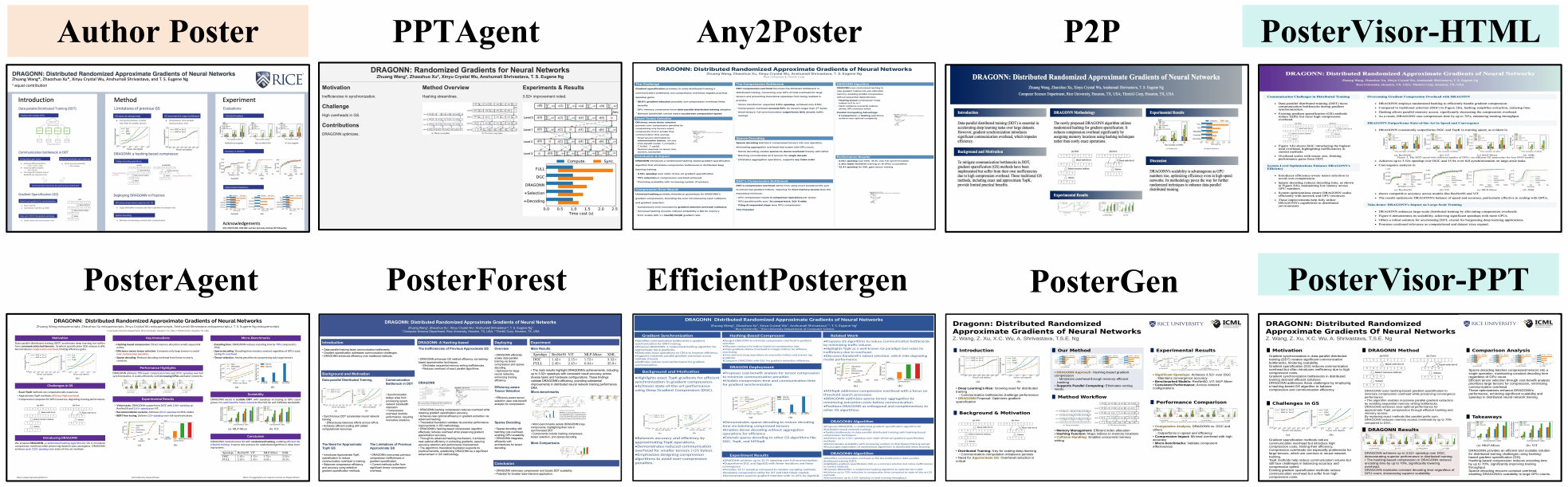}
\caption{Qualitative comparison for one Paper2Poster paper across the author poster, seven automatic baselines, and the two \ourmethod realizations.}
\label{fig:qualitative}
\end{figure*}

Figure~\ref{fig:qualitative} illustrates missing evidence, excessive whitespace, text overload, and undersized visuals in representative baseline outputs. The final outputs use space more fully and foreground key visuals; mechanism evidence comes from the ablation configurations and repair logs below.

On the 30-paper secondary set, \ourmethod-PPT raises PosterGen's VLM Overall from 3.91 to 4.01 and Raw PaperQuiz Overall from 78.9 to 85.5, ranking first among the evaluated automatic methods on both. Full results appear in the supplement.

\subsection{Ablation Study}

We conduct format-specific ablations. For HTML, we compare combinations of SGC, content/structure RCE, and rendered-geometry RCE against the P2P base. For PPTX, we remove one implementation component from the full system at a time. \textbf{Crucially, \textit{Evidence Grounding} and the \textit{Repair Guard} jointly implement the dependency closure ($\Gamma_t$), acting as local constraints that uphold the global persistent contract (SGC).} Specifically, \textit{Evidence Grounding} supplies required visual dependencies, the \textit{Visual Constraint} prevents incompatible large assets from sharing a column, the \textit{Geometry Verifier} checks pre-render spatial feasibility, and the \textit{Repair Guard} prevents local edits from hiding global structural failures.

\begin{table}[!t]
\centering
\normalsize
\setlength{\tabcolsep}{0.8pt}
\begin{tabular*}{\columnwidth}{@{\extracolsep{\fill}}l*{6}{r}@{}}
\toprule
\multirow{2}{*}{Configuration}
& \multicolumn{3}{c}{VLM-as-Judge ($\uparrow$)}
& \multicolumn{3}{c}{Raw PaperQuiz ($\uparrow$)} \\
\cmidrule(lr){2-4} \cmidrule(lr){5-7}
& Aes. & Info. & All & Verb. & Interp. & All \\
\midrule
P2P (base)
& 3.500 & 3.913 & 3.707
& \underline{55.67} & \underline{76.60} & \underline{66.14} \\
\midrule
SGC
& 3.530 & 3.900 & 3.715
& 40.42 & \textbf{77.78} & 59.10 \\
$+$ C/S
& \underline{3.562} & 3.938 & 3.729
& 45.28 & 76.38 & 60.83 \\
$+$ C/S/G
& \textbf{3.657} & \underline{3.980} & \underline{3.818}
& 49.74 & 69.76 & 59.75 \\
Full
& 3.380 & \textbf{4.263} & \textbf{3.822}
& \textbf{58.23} & 75.40 & \textbf{66.82} \\
\bottomrule
\end{tabular*}
\caption{HTML ablations on 100 papers. C, S, and G denote content-, structure-, and geometry-level RCE; ``$+$'' indicates cumulative addition to SGC. Each configuration is generated independently; the base and full-system results match Table~\ref{tab:main_results}.}
\label{tab:ablation}
\end{table}

\begin{figure*}[!t]
\centering
\begin{minipage}[t]{0.49\textwidth}
\centering
\includegraphics[width=\linewidth]{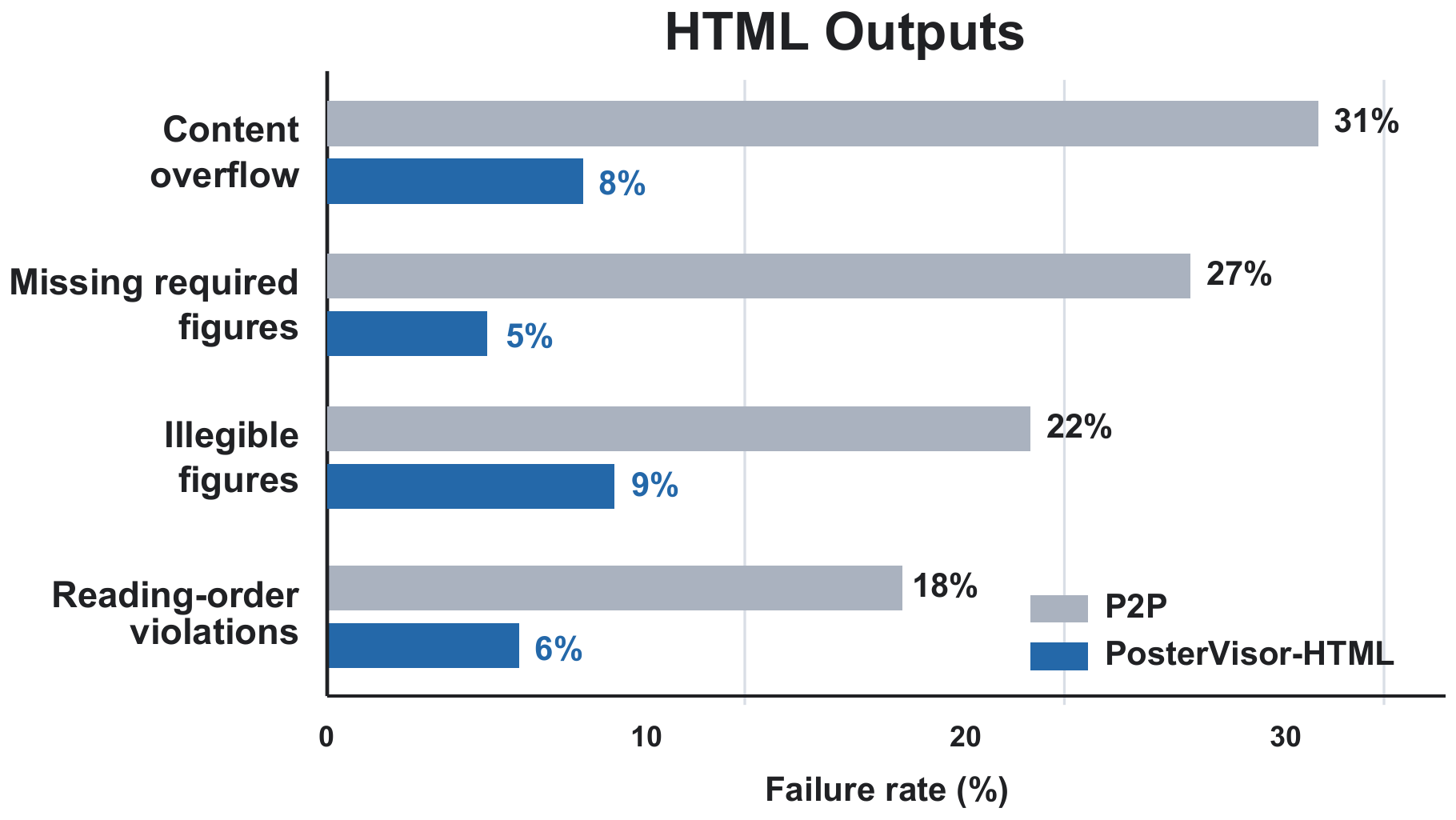}
\captionof{figure}{Output-level failure-signal rates across 100 matched HTML cases: \ourmethod-HTML versus P2P.}
\label{fig:failure}
\end{minipage}
\hfill
\begin{minipage}[t]{0.49\textwidth}
\centering
\includegraphics[width=\linewidth]{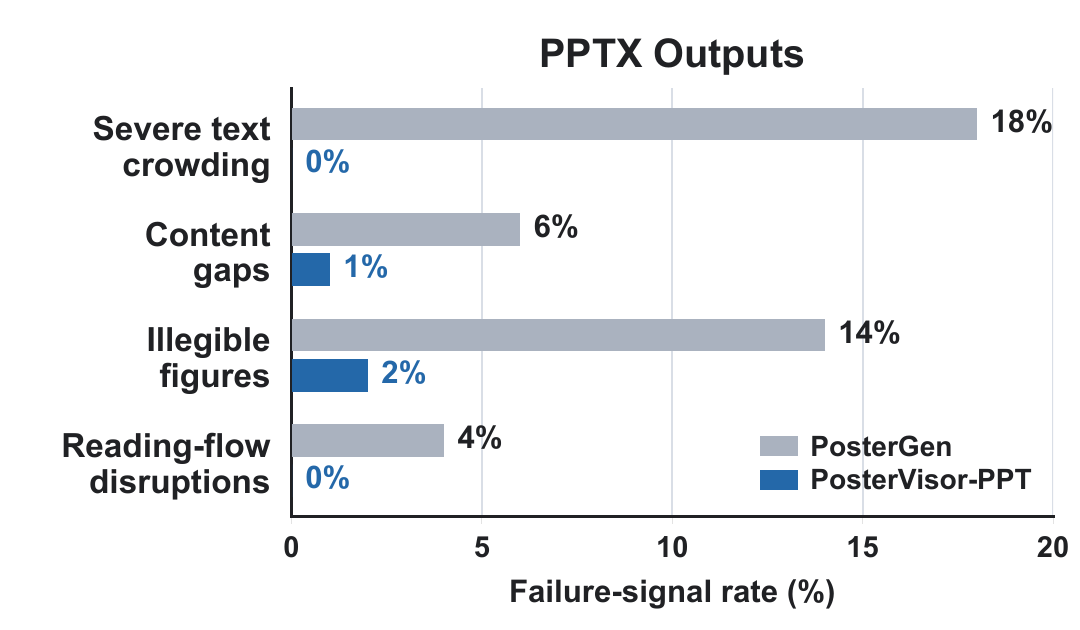}
\captionof{figure}{Output-level failure-signal rates across 100 matched PPTX cases: \ourmethod-PPT versus PosterGen.}
\label{fig:ppt_failure}
\end{minipage}
\end{figure*}

\begin{table}[!t]
\centering
\normalsize
\setlength{\tabcolsep}{2.0pt}
\begin{tabular*}{\columnwidth}{@{\extracolsep{\fill}}l rrrr@{}}
\toprule
Configuration & VLM & QA & Vis. & Struct. \\
& Overall & Overall & demot. & repair \\
\midrule
Full system & \textbf{3.818} & 64.47 & 27 & 217 \\
\midrule
w/o Evidence Grounding & 3.537 & 64.60 & 18 & 162 \\
w/o Visual Constraint & 3.622 & 66.40 & 33 & 179 \\
w/o Geometry Verifier & 3.573 & 66.50 & 0 & 200 \\
w/o Repair Guard & 3.496 & \textbf{67.30} & 29 & 6 \\
\bottomrule
\end{tabular*}
\caption{Diagnostic leave-one-component-out ablations for the PPTX realization. VLM and QA are Overall scores; event columns report totals over each run.}
\label{tab:ppt_loo_main}
\end{table}

Table~\ref{tab:ablation} shows non-monotonic trade-offs among the HTML ablations, with only the full system exceeding P2P on both aggregate objectives. In Table~\ref{tab:ppt_loo_main}, removing any PPTX component lowers VLM Overall from 3.818 to 3.496--3.622. Removing the Geometry Verifier eliminates visual-demotion events, while removing the Repair Guard reduces structural-repair events from 217 to 6, consistent with their intended control roles. The QA changes reveal a content--presentation trade-off rather than uniform gains. Complete switch definitions, additional PPTX configurations, and event counts appear in the supplement.

\subsection{Analysis}

\paragraph{Failure-Mode Reduction.}
Figures~\ref{fig:failure} and~\ref{fig:ppt_failure} show complementary reductions in output-level failure signals: HTML overflow falls from 31\% to 8\% and missing required figures from 27\% to 5\%; PPTX severe text crowding drops from 18\% to 0\% and illegible figures from 14\% to 2\%.

\paragraph{RCE Repair Activity.}
The archived HTML runs contain 159 deterministic local repairs, all followed by rule-based rechecking: 41 are followed by a subsequent full check, while 118 retain only an inline residual check. Regression analysis is restricted to the 41 repair stages with a subsequent full check. Comparing the same rule codes across each adjacent full-check pair yields 75 code-level assertion transitions, of which eight are pass-to-fail transitions across eight different posters. Seven appear only after model-based regeneration following a deterministic patch, while one is already present in the deterministic residual check. Crucially, RCE's post-repair rechecking detects these observed regressions and routes them through the bounded repair loop. Five regressed codes are eliminated before finalization; the remaining three persist after the repair budget is exhausted and are explicitly recorded as residual violations. This directly demonstrates an operational advantage over single-pass checking: subject to its repair budget, RCE prevents observed regressions from propagating silently.

\begin{table}[!t]
\centering
\small
\setlength{\tabcolsep}{2.2pt}
\renewcommand{\arraystretch}{1.08}
\begin{tabular}{@{}p{0.08\columnwidth}p{0.61\columnwidth}p{0.23\columnwidth}@{}}
\toprule
Case & Violation and scoped repair & Recheck \\
\midrule
H01 & Three required visuals missing $\rightarrow$ restore affected placeholders & Content and global checks pass \\
H02 & Required visual lacks two-column span $\rightarrow$ local layout repair and rerender & Violations: $2\!\to\!1\!\to\!0$ \\
U01 & Four-column contract, three-column output $\rightarrow$ two bounded local retries & Unresolved and logged \\
\bottomrule
\end{tabular}
\caption{Selected auditable HTML repair trails. These examples illustrate RCE behavior rather than aggregate repair reliability.}
\label{tab:rce_trails}
\end{table}

\paragraph{Generation Overhead.}
\ourmethod-PPT increases calls from 8.05 to 19.15 and mean latency from 160.32 to 505.49 seconds per poster. Its decomposed requests nevertheless reduce total tokens from 104{,}251 to 67{,}066 and the token-only cost estimate from \$0.297 to \$0.228 per poster.

\subsection{Human Evaluation}

We recruit 15 annotators with AI backgrounds and graduate-level research training: two doctoral researchers and 13 current master's students or master's degree holders. After standardized training, all annotators complete 1{,}175 blind seven-source ranking trials; 13 also complete 2{,}588 blind within-pipeline pairwise trials. In each ranking trial, annotators may select up to three poster sources or abstain at any position. Top-1, Top-2, and Top-3 selections receive $3/2/1$ points, and unselected sources share the mean of the remaining ranks. Pairwise trials compare each \ourmethod realization with its matched base generator, and win rates exclude ties.

\begin{table}[!t]
\centering
\small
\setlength{\tabcolsep}{1.1pt}
\renewcommand{\arraystretch}{0.94}
\begin{tabular*}{\columnwidth}{@{\extracolsep{\fill}}l ccc cc@{}}
\toprule
\multicolumn{6}{l}{\textit{(a) Ranking over seven poster sources}} \\
Method & Top-1 & Top-2 & Top-3 & Score ($\uparrow$) & Rank ($\downarrow$) \\
\midrule
Author Poster & 341 & 198 & 186 & 1605 & 3.18 \\
\midrule
P2P & 164 & \underline{225} & \textbf{209} & \underline{1151} & \underline{3.73} \\
PosterGen & 95 & 128 & \underline{199} & 740 & 4.30 \\
PosterAgent & 59 & 98 & 131 & 504 & 4.68 \\
PPTAgent & 64 & 62 & 55 & 371 & 4.93 \\
\midrule
\ourmethod-PPT & \textbf{254} & \textbf{277} & 183 & \textbf{1499} & \textbf{3.29} \\
\ourmethod-HTML & \underline{184} & 169 & 176 & 1066 & 3.89 \\
\bottomrule
\end{tabular*}

\vspace{0.2ex}

\begin{tabular*}{\columnwidth}{@{\extracolsep{\fill}}l c c@{}}
\toprule
\multicolumn{3}{l}{\textit{(b) Pairwise preference per pipeline}} \\
Comparison & O/B/T & Win \% (95\% CI) \\
\midrule
PPT / PosterGen & 898/341/57 & \textbf{72.5} (61.6--83.4) \\
HTML / P2P & 653/540/99 & 54.7 (45.6--63.8) \\
\bottomrule
\end{tabular*}
\caption{Human evaluation: seven-source ranking (top) and matched-pipeline pairwise outcomes (bottom), with cluster-robust 95\% CIs.}
\label{tab:human_eval}
\end{table}

Table~\ref{tab:human_eval} shows \ourmethod-PPT ranks first among automatic methods and wins 72.5\% of non-tied comparisons with PosterGen (95\% CI, 61.6--83.4\%). \ourmethod-HTML secures more Top-1 selections than P2P but yields a 54.7\% win rate (CI includes 50\%). This statistical tie is expected, as P2P is already highly optimized for aesthetics. Crucially, \ourmethod achieves visual parity while significantly improving factual recoverability (Table~\ref{tab:main_results}). CIs are clustered by annotator and paper~\cite{cameron2011robust}.

\section{Conclusion}

We presented \ourmethod, which coordinates scientific content, visual evidence, and geometry through a persistent SGC and RCE. On the primary benchmark, both realizations obtain higher observed VLM Overall and Raw PaperQuiz Overall than their matched baselines and tie for the highest displayed automatic VLM Overall of 3.82. \ourmethod-PPT additionally gains 5.94 Raw PaperQuiz points and receives 72.5\% of non-tied preferences over PosterGen. These results support persistent contracts, scoped repair, and rechecking as practical control mechanisms for the two pipelines studied here.

\bibliography{aaai2027}



\end{document}


\maketitle

\renewcommand{\thefigure}{S\arabic{figure}}
\renewcommand{\thetable}{S\arabic{table}}

\section{Operationalizing the SGC and RCE}
The two implementations specialize the frozen contract defined in the main
paper through format-specific states, validators, and bounded repair operators.

\subsection{Worked Archived SGC Instance}
\label{sec:worked-sgc}
This section provides an implementation-level example of a frozen SGC and its
checkpoint evidence. The archived HTML record binds a warm-start claim to a
required source visual, its planned placement, executable survival checks, and
readability guidance. The assignments and validator bindings are read directly
from the saved contract and repair trail, without additional model inference or
post-hoc reclassification.

\begin{table*}[t]
\centering
\small
\setlength{\tabcolsep}{4pt}
\begin{tabular}{@{}p{0.17\textwidth}p{0.28\textwidth}p{0.22\textwidth}p{0.25\textwidth}@{}}
\toprule
Grounded requirement & Saved contract realization & Operational class & Validator and archived evidence \\
\midrule
Warm-start claim and source visual &
$K_c$: claim-bearing section and required source visual; $K_l$: assigned column
and multi-column span &
$K_c/K_l$ contract fields; the linked requirement is executable only with a
validator binding &
Fields are retained in the frozen contract and supplied to content and layout
generation \\
\addlinespace
Required visual survives writing &
The required visual must appear as a valid numeric Markdown placeholder after
content generation &
$\mathcal{E}$: executable content assertion &
\url{MISSING_FIGURE_PLACEHOLDER}; failed before repair and passed after
deterministic reference restoration \\
\addlinespace
Required visual survives layout &
Editable HTML must retain the corresponding image reference before source-asset
embedding &
$\mathcal{E}$: executable layout assertion &
\texttt{MISSING\_VISUAL\_REF}; passed at the archived layout checkpoint \\
\addlinespace
Readable, logically placed visual &
Keep trends and annotations visible; place figures logically to support the
storyline &
$\mathcal{S}$: soft guidance &
Retained as generation guidance unless compilation produces a supported
measurable predicate \\
\bottomrule
\end{tabular}
\caption{One archived paper-specific SGC instance. The
$\mathcal{E}/\mathcal{S}$ assignments and their validator bindings are read from
the saved frozen-contract record; checkpoint artifacts provide the observed
pass/fail evidence.}
\label{tab:worked-sgc}
\end{table*}

The initial content omitted the required visual. Deterministic repair restored
its reference, and the residual recheck passed. The editable-HTML checkpoint
then raised \url{COLUMN_COUNT_DRIFT}; a bounded layout retry restored the frozen
column target, and the final checkpoint-wide recheck passed. The case therefore
illustrates both executable constraints in $\mathcal{E}$ and soft preferences in
$\mathcal{S}$ without treating every stored plan field as a compliance test.

\section{Format-Specific Instantiations}
The HTML \texttt{GlobalPlan} and PPTX blueprint become the format-specific
frozen SGC representations only after normalization and assertion compilation
have instantiated $K_c$, $K_l$, $\mathcal{E}$, and $\mathcal{S}$. Before that
point, they are candidate plans or blueprints; the resulting complete contract
is then frozen and serialized as $K$.

\subsection{\ourmethod-HTML}
\label{sec:html}

The \ourmethod-HTML runs on the main 100-paper Paper2Poster
benchmark~\cite{pang2025paper2poster} use GPT-4o for generation and visual
analysis. The frozen plan uses a 650-word poster
budget, balanced density, and at least three required visual assets.

\subsubsection{HTML Contract Schema}

The GlobalPlan links section identifiers, sources, budgets, and positions to
visual roles, spans, size constraints, readability guidance, and grounded
rubric records.

The planner grounds claim-oriented sections and visual roles in extracted text,
captions, and image geometry. It distributes the text budget, prioritizes the
key visual, and emits schema-valid JSON; dense tables may be represented as
result cards or simplified tables.

\subsubsection{Pre-Render Content Verification}

The content checkpoint verifies required sections and visual placeholders, the
650-word poster budget, and a per-section limit of 1.5 times the planned budget.
Section-count drift is blocking only when a required section is missing;
readability is deferred to rendered-output inspection.

Deterministic repair restores planned order, required visual references, and
budget compliance. A missing section triggers a targeted retry restricted to
the implicated section, followed by the same content checks.

\subsubsection{Structure and Rendered-Geometry Verification}

The editable-layout checkpoint checks required image references, the key
visual's intended span, and \url{COLUMN_COUNT_DRIFT}. DOM repair can restore
references or span metadata; column drift triggers a bounded layout retry and
the same structural recheck.

Playwright then measures rendered bounding boxes. Blocking assertions cover
overlap above 2\% of the smaller element, overflow beyond 8 pixels, aspect-ratio
distortion above 10\%, text columns below 180 pixels or 12\% of poster width,
within-row width ratios above $4{:}1$, content-width coverage below 70\%, section
blank ratios above 35\%, and figure containers below 15\% of their column width.
Supported figure-size predicates are enforced only when compiled into
$\mathcal{E}$; other readability preferences remain in $\mathcal{S}$.

Content and editable-layout verification each permit up to two repair rounds;
the subsequent layout--render--check loop permits at most three attempts along
the same trajectory. A retained artifact may be returned after budget
exhaustion, but any unresolved executable assertion keeps compliance false.

\subsection{\ourmethod-PPT}
\label{sec:ppt}
The PPTX implementation normalizes the candidate blueprint before freezing and
then verifies content, editable geometry, typography, and rendered output.

\subsubsection{Blueprint Contract}

The Evidence Reader supplies indexed paper sections and a scored visual
inventory. The Orchestrator uses the highest-ranked candidates to construct a
five-to-eight-panel blueprint with grounded claims, sources, budgets, and visual
bindings.

Before normalization and assertion compilation, the planner output is only a
candidate blueprint. The finalized blueprint binds panel claims, evidence,
budgets, and layout priorities to the compiled $\mathcal{E}/\mathcal{S}$ records
and is then frozen as the PPTX-specific SGC.

During normalization, the Orchestrator designates exactly one high-importance
anchor panel, places it at the top of the middle column, and binds it to the key
visual. The PPTX format policy clamps per-panel budgets to 50--140 words and
normalizes them toward a poster-wide target of approximately 400 body words.
Before the SGC is frozen, required visual and source identifiers are resolved
against the supplied inventories; unresolved identifier or anchor conflicts
remain plan violations.

\subsubsection{Adaptive Global Geometry}

During SGC normalization, the Orchestrator selects a near-equal global column
profile from panel roles, estimated load, and visual evidence. The resulting
allocations instantiate $K_l$, and panel budgets in $K_c$ are rescaled by the
assigned column width before freezing.

The \textit{Geometry Verifier} estimates column coverage from text budgets,
typography, padding, and visual aspect ratios, using an accepted band of
0.75--1.10. For an overfull candidate, bounded normalization may move one
non-anchor panel, replace one non-protected visual binding with grounded textual
evidence, or split one long text panel. Feasibility is recomputed after each
action; unresolved conflicts remain plan violations.

During the same normalization stage, the PPTX format policy imposes a
deterministic large-visual allocation constraint: no column may contain two
large figures or a large figure together with a large table. For this
constraint, a figure is considered large when its estimated height exceeds
45\% of the column height, and a table is considered large when its estimated
height exceeds 35\%. The key visual and any visual bound to the anchor panel
cannot be converted into text-only evidence requirements. If the allowed
candidate-contract normalization actions cannot resolve the allocation
conflict, it remains a plan violation.

\subsubsection{Plan/Content Verification and Repair}

After the Content Agent produces the poster content $C$, the content checkpoint
compares the storyboard with the frozen SGC blueprint. Deterministic predicates
check panel presence (\url{SECTION_PRESENT}), title identity
(\url{SECTION_TITLE}), required visual evidence (\url{REQUIRED_FIGURE}), and
panel and poster-wide word budgets. Two task-scoped model predicates evaluate
the semantic records already compiled into $\mathcal{E}$:
\url{UNCOVERED_CLAIM} tests coverage of each panel's \texttt{main\_claim} and
\texttt{must\_mention} facts, while \url{OUT_OF_SOURCE_EVIDENCE} tests whether
the stated evidence is consistent with the panel's frozen
\texttt{source\_sections}. Both codes are blocking executable assertions. A
model-call failure returns $\bot$ and leaves the assertion unresolved rather
than passing it.

Deterministic repairs restore missing panels or titles, insert required figures,
and truncate text to the applicable budget. The two semantic violation codes
permit only a \texttt{scoped\_content\_patch}: the implicated panel is rewritten
to cover its frozen claim and required facts using only its allowed source
sections. Each applied patch is followed by the same content checks. The format
policy allows up to three scoped content-repair calls and stops when verification
no longer improves.

\subsubsection{Layout Verification and Typed Repair}

At the editable-layout checkpoint before rendering, the Checker evaluates
executable assertions in $\mathcal{E}$ for page coverage, per-column coverage,
column imbalance, bottom whitespace, key-visual and main-result size,
aspect-ratio distortion, and overlap. In the reported PPTX format policy, the
instantiated blocking bounds are 0.82--0.94 for page coverage, 0.78--0.96 for
column coverage, at most 0.18 for column imbalance, and at most 0.12 for the
bottom-blank ratio. After typography is applied, the post-typography layout
checkpoint remeasures text bounds and collisions and enforces a minimum
body-font size of 36 points. Failed assertions are routed through bounded RCE
and must pass before the artifact can be certified contract-compliant.

At the rendered-output checkpoint, the Checker applies full-poster task-scoped
VLM visual-QA together with rule-first panel- and column-level verification.
The rule-based path recomputes deterministic geometry from the rendered styled
layout, while task-scoped VLM predicates evaluate already compiled assertions
in $\mathcal{E}$, including applicable figure--text consistency and numerical-
evidence requirements. A model finding affects contract compliance only when
it evaluates an assertion in $\mathcal{E}$; other critic observations remain
diagnostic. The configured severity threshold determines which actionable
violations are routed to repair, but it does not convert a failed executable
assertion into a pass. A model-call failure returns $\bot$ and is never treated
as verified compliance.

At rendered-output RCE, the repair router selects a typed operator from
$\mathcal{U}_i$, such as resizing or moving a visual within its column,
adjusting text, adding compact evidence, or dropping a non-key visual. The
dependency closure is implemented jointly by Evidence Grounding and the
\textit{Repair Guard}: Evidence Grounding supplies required-visual dependencies,
while the Repair Guard enforces the resulting repair scope by validating targets
and parameters, preserving the canvas and frozen column assignments, and
preventing removal of the key visual. Format-wide readability bounds remain
enforced, including minimum scales of 0.65 for ordinary figures and 0.75 for
tables and the key visual. Up to three repair--rerender--recheck iterations are
allowed; exhausted budgets return the latest artifact with unresolved assertions
recorded in $L$ and compliance kept false.

\section{Mechanism Validation}
\label{sec:supp-mechanism}
This section audits the archived generation-time logs and artifacts from the
full \ourmethod-HTML and \ourmethod-PPT GPT-4o runs underlying their rows in the
main 100-paper comparison. It invokes no additional model inference and uses no
evaluation-model logs.

\subsection{HTML Recheck and Regression Audit}
\label{sec:html-mechanism-audit}

At the immediate post-repair check, all eight detected pass-to-fail regression
transitions were recorded as \textsc{fail} rather than $\bot$. Five were
subsequently resolved before finalization, whereas three remained as residual
violations after the repair budget was exhausted.

\subsection{PPTX Mechanism Audit}
\label{sec:pptx-mechanism-audit}

\paragraph{SGC construction boundary.}
The audit observes 21 candidate-blueprint geometry mutations across 18 posters.
They occur before the blueprint is frozen and are therefore SGC construction,
not post-freeze RCE repair.

\begin{table}[!t]
\centering
\small
\setlength{\tabcolsep}{4pt}
\renewcommand{\arraystretch}{1.08}
\begin{tabular}{@{}>{\raggedright\arraybackslash}p{0.35\columnwidth}
                    r
                    >{\raggedright\arraybackslash}p{0.42\columnwidth}@{}}
\toprule
Control stage & Events & Mechanism role \\
\midrule
Candidate-blueprint normalization & 21 & Pre-freeze SGC construction \\
Content-checkpoint RCE & 153 & Content repair and recheck \\
Editable-layout RCE & 121 & Geometry repair and recheck \\
Post-typography RCE & 24 & Styled-geometry repair and recheck \\
Rendered-output RCE & 103 & Guarded repair, rerendering, and recheck \\
\midrule
Total & 422 & \\
\bottomrule
\end{tabular}
\caption{PPTX control events observed in the main-benchmark GPT-4o
\ourmethod-PPT runs over 100 papers. Counts are per-event repair records, not
violation counts or state transitions; one record may bundle multiple violation
codes. The 21 pre-freeze blueprint-geometry mutations are excluded from the 401
post-freeze records by definition.}
\label{tab:pptx-audit-actions}
\end{table}

Audit recheck coverage is 1.0 for content, editable-layout, and post-typography
repair records. Recheck coverage for the rendered-output critic loop is also
1.0 (29/29 critic-loop repair trajectories): all 29 posters whose critic loop
applied repairs completed the required post-repair recheck. Each trajectory
completed violation detection, repair writing, repair application,
rerendering, and full-poster rechecking. The denominator is poster-level
critic-loop repair trajectories, not the 103 rendered-output event records in
Table~\ref{tab:pptx-audit-actions}; rechecking is performed over the full poster
state rather than paired to individual violation codes. Thus, no repaired
trajectory inherits a preceding check result without revalidation.
The reported repair counts therefore use distinct scopes. The main paper's 159
events are deterministic local repairs from the HTML pipeline;
Table~\ref{tab:pptx-audit-actions} counts 422 PPTX control records across
checkpoints, including 21 pre-freeze mutations; and the 217 value in main-paper
Table~3 is a PPTX category-specific structural-repair counter. These quantities
do not share a denominator and should not be compared as alternative totals.
\section{Ablation Study and Visual Evidence}
For the main 100-paper GPT-4o benchmark, this section reports the complete PPTX
progressive study and an expanded decomposition of the
leave-one-component-out results summarized in Table~3 of the main paper.
The quantitative HTML ablation table is not repeated here.

\subsection{PPTX Progressive Configurations}

The progressive study begins with the PosterGen
baseline~\cite{zhang2025postergen}. L2 adds only a blueprint scaffold, not the
finalized blueprint used as the SGC. L3 adds
source-grounded claims, evidence anchors, and required-visual dependencies;
after normalization and assertion compilation, the resulting blueprint is
frozen as the grounded SGC. L4 activates the
pre-render Visual Constraint, Geometry Verifier, and Repair Guard, and L5 adds
rendered-output RCE. Tables~\ref{tab:ppt-config-full}
and~\ref{tab:ppt-progressive-full} report the switches and results.

\begin{table}[t]
\centering
\footnotesize
\setlength{\tabcolsep}{1.5pt}
\begin{tabular}{l l ccccc}
\toprule
Layer & Added & EG & VC & GV & RG & RR \\
\midrule
L1 PosterGen & baseline & -- & -- & -- & -- & -- \\
L2 blueprint scaffold & scaffold & $\times$ & $\times$ & $\times$ & $\times$ & $\times$ \\
L3 grounded SGC & EG & \checkmark & $\times$ & $\times$ & $\times$ & $\times$ \\
L4 pre-render & VC/GV/RG & \checkmark & \checkmark & \checkmark & \checkmark & $\times$ \\
L5 full & rendered RCE & \checkmark & \checkmark & \checkmark & \checkmark & \checkmark \\
\bottomrule
\end{tabular}
\caption{Switches in the progressive \ourmethod-PPT study. EG: Evidence
Grounding; VC: Visual Constraint; GV: Geometry Verifier, including pre-render
layout verification; RG: Repair Guard; RR: rendered-output RCE.}
\label{tab:ppt-config-full}
\end{table}

\begin{table*}[t]
\centering
\small
\setlength{\tabcolsep}{5pt}
\begin{tabular}{@{}l ccc ccc@{}}
\toprule
\multirow{2}{*}{Configuration}
& \multicolumn{3}{c}{VLM-as-Judge ($\uparrow$)}
& \multicolumn{3}{c}{Raw PaperQuiz Accuracy ($\uparrow$)} \\
\cmidrule(lr){2-4} \cmidrule(lr){5-7}
& Aesth. & Info. & Overall & Verbatim & Interpretive & Overall \\
\midrule
L1 PosterGen & 3.130 & 3.800 & 3.460 & 43.90 & 73.15 & 58.53 \\
L2 blueprint scaffold & 3.040 & 4.100 & 3.570 & 45.17 & 76.10 & 60.64 \\
L3 + grounded SGC & 3.060 & 4.030 & 3.540 & 48.91 & 75.49 & 62.20 \\
L4 + pre-render controls & 3.100 & 4.050 & 3.580 & 47.09 & 74.70 & 60.89 \\
L5 full \ourmethod-PPT & 3.607 & 4.030 & 3.818 & 53.02 & 75.91 & 64.47 \\
\bottomrule
\end{tabular}
\caption{Complete progressive \ourmethod-PPT results on the main GPT-4o
benchmark. Because configurations are independently generated, adjacent rows
describe configuration-level transitions rather than isolated marginal effects.}
\label{tab:ppt-progressive-full}
\end{table*}

\makeatletter
\setlength{\@dblfptop}{0pt}
\setlength{\@dblfpsep}{12pt}
\setlength{\@dblfpbot}{0pt plus 1fil}
\makeatother
\begin{table*}[!t]
\centering
\footnotesize
\begin{minipage}[t]{0.56\textwidth}
\centering
\textit{(a) Leave-one-out quality and information retention}\par\smallskip
\setlength{\tabcolsep}{2.5pt}
\begin{tabular}{@{}l ccc ccc@{}}
\toprule
\multirow{2}{*}{Configuration} & \multicolumn{3}{c}{VLM-as-Judge} & \multicolumn{3}{c}{Raw PaperQuiz Accuracy} \\
\cmidrule(lr){2-4} \cmidrule(lr){5-7}
& Aesth. & Info. & Overall & Verbatim & Interpretive & Overall \\
\midrule
full & 3.607 & 4.030 & 3.818 & 53.02 & 75.91 & 64.47 \\
\midrule
w/o Visual Constraint & 3.080 & 4.163 & 3.622 & 60.80 & 72.00 & 66.40 \\
w/o Geometry Verifier & 3.057 & 4.090 & 3.573 & 61.50 & 71.50 & 66.50 \\
w/o Evidence Grounding & 3.057 & 4.017 & 3.537 & 58.80 & 70.30 & 64.60 \\
w/o Repair Guard & 3.016 & 3.975 & 3.496 & 62.40 & 72.20 & 67.30 \\
\bottomrule
\end{tabular}
\end{minipage}\hfill
\begin{minipage}[t]{0.40\textwidth}
\centering
\textit{(b) Recorded mechanism events}\par\smallskip
\setlength{\tabcolsep}{2.5pt}
\begin{tabular}{@{}l ccc@{}}
\toprule
Config. & \shortstack{Active geom.\\reports} & \shortstack{Vis.\\demot.} & \shortstack{Struct.\\repairs} \\
\midrule
\multicolumn{4}{@{}l}{\textit{Leave-one-out configurations}} \\
full & 100 & 27 & 217 \\
w/o GV & 0 & 0 & 200 \\
w/o VC & 100 & 33 & 179 \\
w/o RG & 100 & 29 & 6 \\
w/o EG & 100 & 18 & 162 \\
\midrule
\multicolumn{4}{@{}l}{\textit{Progressive-stage controls}} \\
L2 blueprint scaffold & 0 & 0 & 0 \\
L3 grounded SGC & 0 & 0 & 0 \\
L4 pre-render controls & 100 & 29 & 112 \\
\bottomrule
\end{tabular}
\end{minipage}
\caption{Diagnostic \ourmethod-PPT leave-one-component-out configurations and
recorded control events. VLM quality and Raw PaperQuiz Accuracy are aggregate
results for each
configuration. Whenever the geometry-gate node is instantiated, it emits a
report file; when GV is disabled in the leave-one-out ablation, that file is a
\texttt{status="disabled"} stub with no actions. Active geometry reports
therefore count only
\texttt{status="ok"} records, rather than files. In the focal full versus w/o GV
comparison, the active gate produced 21 geometry mutations across 18 posters,
versus zero mutations when GV was disabled. Visual demotions and structural
repairs are event totals rather than poster counts or repair success rates. EG,
VC, GV, and RG follow the component names defined in
Table~\ref{tab:ppt-config-full}.}
\label{tab:ppt-loo-full}
\end{table*}

\subsection{Same-Paper Visual Ablations}

\begin{figure*}[!t]
\centering
\includegraphics[width=0.86\textwidth]{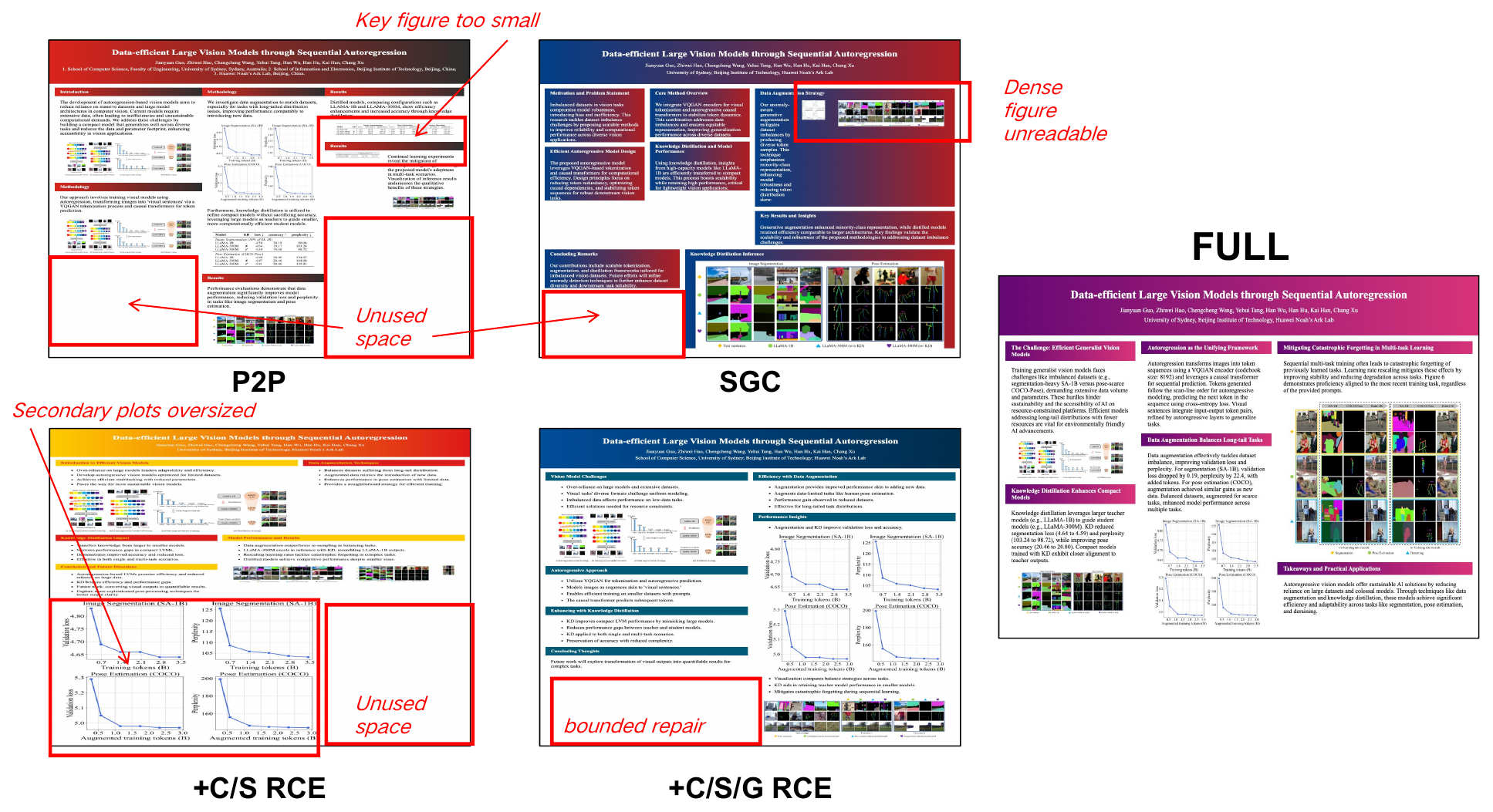}
\par\smallskip
\textbf{(a) Cumulative HTML configurations}

\medskip
\includegraphics[width=0.86\textwidth]{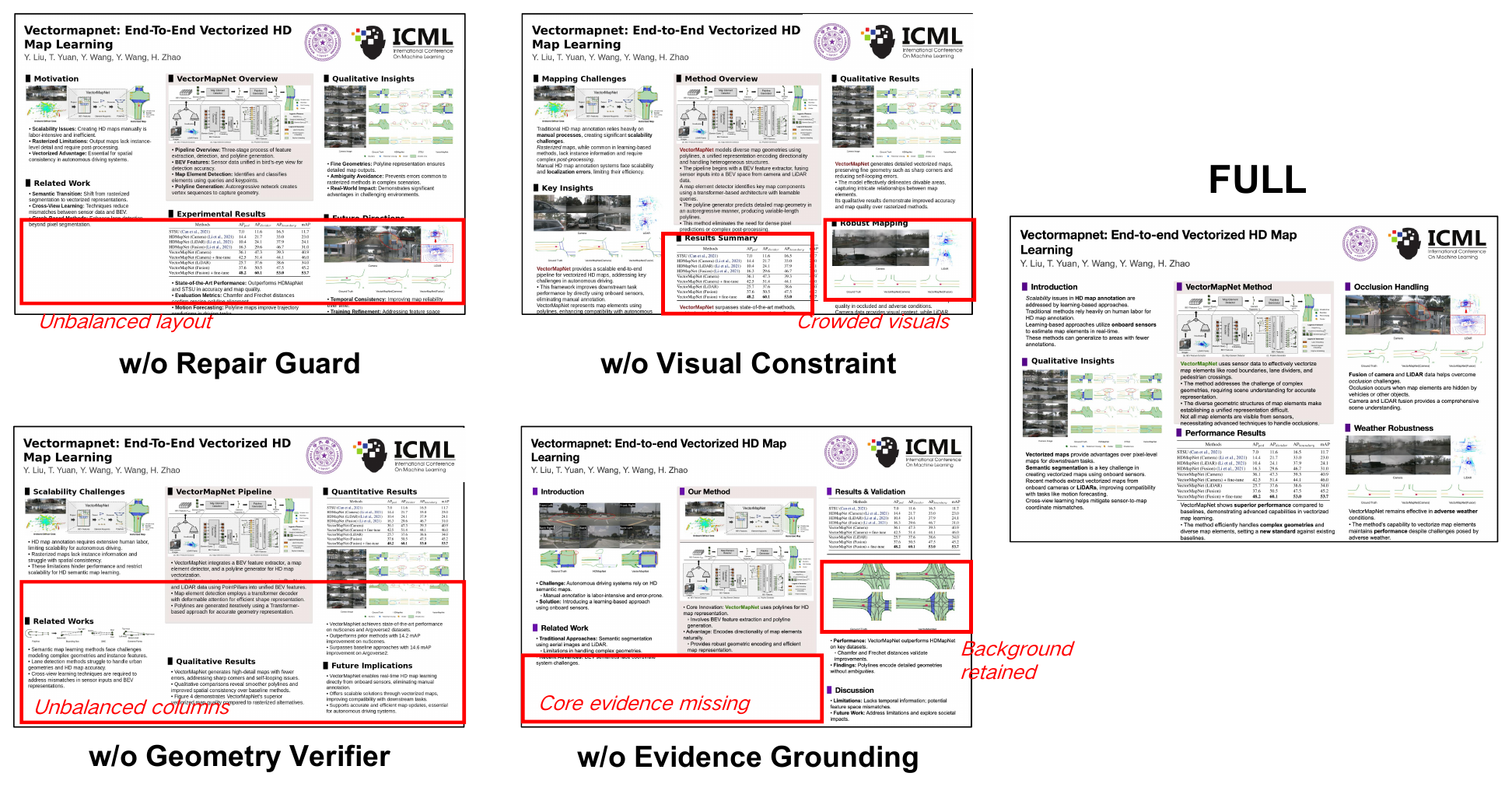}
\par\smallskip
\textbf{(b) Diagnostic PPTX leave-one-out configurations}
\caption{Same-paper visual ablations. In (a), C, S, and G denote content-,
structure-, and rendered-geometry-level RCE, respectively. The $+$C/S/G
configuration adds all three RCE levels to SGC; Full additionally enables
visual analysis, readability-aware figure-size enforcement, and bounded
layout--render--check retries with layout compaction, all performed under the
frozen SGC. Callouts mark undersized or unreadable figures, unused space,
crowded layouts, column imbalance, and misplaced visual priority across the
HTML and PPTX configurations.}
\label{fig:visual-ablation}
\end{figure*}

Figure~\ref{fig:visual-ablation} shows same-paper outputs for the cumulative
HTML configurations and diagnostic PPTX leave-one-out configurations.

\section{Extended Effectiveness}
\label{sec:supp-extended}
\subsection{Cross-Model Robustness}
\label{sec:cross-model}

Table~\ref{tab:cross-model-results} reports the six systems under four
additional model settings. Within each setting, the named model handles every
model-based call in the generation and internal-control pipeline, including
figure scoring and visual analysis, Orchestrator and blueprint planning,
Content Agent and section writing, internal semantic and rendered-poster
critics, and Repair Writer. Thus, the Qwen3.7-Plus setting uses Qwen3.7-Plus
throughout this pipeline, and the same rule applies to GPT-5.5,
Gemini-3.1-Pro-Preview, and Claude-Opus-4.8. The six-system subset was fixed
when these experiments were run; baselines added later to the main comparison
were omitted because their public implementations were unavailable or not
reproducible at that time. GPT-4o does not participate in poster generation,
internal checking, or repair. It is used only during external evaluation, both
as the VLM-as-Judge and as one member of the common PaperQuiz reader ensemble.

\begin{table*}[!t]
\centering
\footnotesize
\renewcommand{\arraystretch}{0.94}
\scriptsize
\setlength{\tabcolsep}{1.2pt}
\resizebox{\textwidth}{!}{%
\begin{tabular}{@{}l*{15}{c}@{}}
\toprule
\multirow{3}{*}{Method}
& \multicolumn{9}{c}{VLM-as-Judge ($\uparrow$)}
& \multicolumn{6}{c}{PaperQuiz ($\uparrow$)} \\
\cmidrule(lr){2-10} \cmidrule(lr){11-16}
& \multicolumn{4}{c}{Aesthetic}
& \multicolumn{4}{c}{Information}
& \multirow{2}{*}{Overall}
& \multicolumn{3}{c}{Raw PaperQuiz Accuracy}
& \multicolumn{3}{c}{Density-Augmented PaperQuiz} \\
\cmidrule(lr){2-5} \cmidrule(lr){6-9}
\cmidrule(lr){11-13} \cmidrule(lr){14-16}
& Elem. & Layout & Engage. & Avg.
& Clarity & Content & Logic & Avg. &
& Verbatim & Interpretive & Overall
& Verbatim & Interpretive & Overall \\
\midrule
\multicolumn{16}{@{}l}{\textit{Qwen3.7-Plus}} \\
GT Poster & 4.07 & 3.90 & 2.70 & 3.56 & 4.09 & 3.96 & 3.89 & 3.98 & 3.77 & 68.25 & 74.73 & 71.49 & 127.75 & 140.29 & 134.02 \\
\midrule
PosterAgent & 2.92 & 3.01 & 1.76 & 2.56 & 2.07 & 3.89 & 4.90 & 3.62 & 3.09 & 60.53 & 72.91 & 66.72 & 114.21 & 137.90 & 126.06 \\
PosterGen & 3.53 & \underline{3.62} & 2.38 & 3.18 & \textbf{4.71} & 3.90 & 4.94 & \underline{4.52} & 3.85 & 53.53 & 72.97 & 63.25 & 107.07 & 145.93 & 126.50 \\
PPTAgent & \textbf{3.75} & 3.45 & \underline{2.43} & \underline{3.21} & 4.61 & 4.02 & 4.91 & 4.51 & \underline{3.86} & 63.48 & \underline{76.45} & 69.97 & 126.96 & \underline{152.91} & 139.93 \\
P2P & 2.90 & 3.00 & 1.93 & 2.61 & 4.35 & \textbf{4.18} & \underline{4.98} & 4.50 & 3.56 & \underline{66.03} & 76.31 & \underline{71.17} & \underline{131.11} & 151.56 & \underline{141.33} \\
\ourmethod-PPT & \underline{3.72} & \textbf{3.89} & \textbf{2.87} & \textbf{3.49} & \underline{4.66} & \underline{4.09} & 4.97 & \textbf{4.57} & \textbf{4.03} & 55.64 & 74.87 & 65.26 & 110.79 & 149.08 & 129.94 \\
\ourmethod-HTML & 3.20 & 3.37 & 2.14 & 2.90 & 4.53 & 3.87 & \textbf{4.99} & 4.46 & 3.68 & \textbf{71.83} & \textbf{77.12} & \textbf{74.47} & \textbf{143.18} & \textbf{153.77} & \textbf{148.47} \\
\midrule
\multicolumn{16}{@{}l}{\textit{GPT-5.5}} \\
GT Poster & 4.07 & 3.90 & 2.70 & 3.56 & 4.09 & 3.96 & 3.89 & 3.98 & 3.77 & 68.25 & 74.73 & 71.49 & 127.75 & 140.29 & 134.02 \\
\midrule
PosterAgent & 2.51 & 2.37 & 2.06 & 2.31 & 2.19 & 3.48 & 3.90 & 3.19 & 2.75 & 67.75 & 75.54 & 71.65 & \underline{125.69} & 140.68 & 133.19 \\
PosterGen & \underline{3.06} & 2.99 & 2.62 & 2.89 & \textbf{3.75} & 3.29 & 3.89 & 3.64 & \underline{3.27} & 56.45 & 74.21 & 65.33 & 112.80 & 148.25 & 130.52 \\
PPTAgent & 2.94 & \textbf{3.12} & \textbf{2.92} & \textbf{2.99} & 3.52 & 2.96 & 3.85 & 3.44 & 3.22 & 48.75 & 75.55 & 62.15 & 97.49 & \underline{151.11} & 124.30 \\
P2P & 2.63 & 2.96 & 2.01 & 2.53 & 3.44 & \textbf{3.99} & \textbf{4.00} & \textbf{3.81} & 3.17 & \underline{73.75} & \underline{77.15} & \underline{75.45} & 124.82 & 130.43 & 127.62 \\
\ourmethod-PPT & \textbf{3.13} & 2.99 & \underline{2.63} & \underline{2.92} & \underline{3.58} & 3.45 & 3.89 & 3.64 & \textbf{3.28} & 60.02 & 75.58 & 67.80 & 119.48 & 150.42 & \underline{134.95} \\
\ourmethod-HTML & 2.85 & \underline{3.02} & 2.20 & 2.69 & 3.48 & \underline{3.78} & \underline{3.97} & \underline{3.74} & 3.22 & \textbf{75.60} & \textbf{79.19} & \textbf{77.39} & \textbf{149.11} & \textbf{156.20} & \textbf{152.66} \\
\midrule
\multicolumn{16}{@{}l}{\textit{Gemini-3.1-Pro-Preview}} \\
GT Poster & 4.07 & 3.90 & 2.70 & 3.56 & 4.09 & 3.96 & 3.89 & 3.98 & 3.77 & 68.25 & 74.73 & 71.49 & 127.75 & 140.29 & 134.02 \\
\midrule
PosterAgent & 2.66 & 2.46 & 2.53 & 2.55 & 2.29 & 3.19 & \textbf{4.45} & 3.31 & 2.93 & \underline{65.07} & \underline{76.17} & \underline{70.62} & 123.50 & 144.95 & 134.22 \\
PosterGen & 2.62 & 2.56 & \textbf{2.58} & \textbf{2.59} & 3.73 & \underline{3.40} & 4.10 & \underline{3.74} & \textbf{3.17} & 55.67 & 74.61 & 65.14 & 111.23 & 149.09 & 130.16 \\
PPTAgent & \textbf{2.85} & \underline{2.64} & 2.17 & 2.55 & 2.88 & 2.08 & 4.23 & 3.06 & 2.81 & 32.66 & 73.14 & 52.90 & 65.32 & 146.28 & 105.80 \\
P2P & 2.16 & 2.26 & 2.09 & 2.17 & 3.62 & \textbf{3.57} & \underline{4.35} & \textbf{3.85} & 3.01 & 55.96 & 75.06 & 65.51 & 110.17 & 147.80 & 128.99 \\
\ourmethod-PPT & \underline{2.72} & 2.42 & \underline{2.55} & \underline{2.56} & \underline{3.74} & 3.24 & 4.02 & 3.67 & \underline{3.12} & \textbf{71.11} & \textbf{78.12} & \textbf{74.62} & \textbf{142.15} & \textbf{156.17} & \textbf{149.16} \\
\ourmethod-HTML & 2.40 & \textbf{2.73} & 2.29 & 2.47 & \textbf{3.87} & 3.09 & 4.18 & 3.71 & 3.09 & 63.03 & 75.79 & 69.41 & \underline{125.99} & \underline{151.53} & \underline{138.76} \\
\midrule
\multicolumn{16}{@{}l}{\textit{Claude-Opus-4.8}} \\
GT Poster & 4.07 & 3.90 & 2.70 & 3.56 & 4.09 & 3.96 & 3.89 & 3.98 & 3.77 & 68.25 & 74.73 & 71.49 & 127.75 & 140.29 & 134.02 \\
\midrule
PosterAgent & 2.48 & 2.86 & 2.20 & 2.51 & 2.86 & 3.69 & 3.98 & 3.51 & 3.01 & 70.49 & \underline{77.50} & 74.00 & \underline{130.27} & 143.46 & 136.87 \\
PosterGen & \underline{2.80} & 2.97 & 2.59 & 2.79 & 3.93 & 3.75 & 3.98 & 3.89 & 3.34 & 65.05 & 76.33 & 70.69 & 130.05 & \underline{152.59} & \underline{141.32} \\
PPTAgent & \textbf{2.91} & \textbf{3.43} & \textbf{3.00} & \textbf{3.11} & \textbf{4.01} & 3.60 & \textbf{4.00} & 3.87 & \textbf{3.49} & 58.06 & 75.98 & 67.02 & 116.12 & 151.96 & 134.04 \\
P2P & 2.50 & \underline{3.20} & 2.58 & 2.76 & \underline{4.00} & \underline{3.78} & 3.85 & 3.88 & 3.32 & \underline{72.68} & 77.14 & \underline{74.91} & 128.39 & 136.66 & 132.53 \\
\ourmethod-PPT & 2.70 & 3.12 & \underline{2.81} & \underline{2.88} & \underline{4.00} & \textbf{3.90} & \underline{3.99} & \textbf{3.96} & \underline{3.42} & 63.58 & 76.51 & 70.04 & 126.72 & 152.48 & 139.60 \\
\ourmethod-HTML & 2.44 & 3.17 & 2.63 & 2.75 & 3.96 & \textbf{3.90} & 3.91 & \underline{3.92} & 3.34 & \textbf{74.79} & \textbf{78.14} & \textbf{76.47} & \textbf{146.77} & \textbf{153.35} & \textbf{150.06} \\
\bottomrule
\end{tabular}
}
\caption{Cross-model robustness results. In each setting, the named model is
used for every model-based call in the generation and internal-control
pipeline. GPT-4o does not participate in poster generation, internal checking,
or repair; during external evaluation, it serves as the VLM-as-Judge and as one
member of the common PaperQuiz reader ensemble. The complete GT Poster row
reuses the main-comparison reference values. Bold
and underlined values denote the best and
second-best automatic methods within each setting; rounded ties share the same
mark.}
\label{tab:cross-model-results}
\end{table*}

\subsection{Failure-Signal Audit}

For the HTML comparison in Figure~3 of the main paper, each percentage is
computed over the final rendered posters. The four categories capture
output-level signals of content overflow, missing required figures, illegible
figures, and reading-order disruption. These signals are first identified from
poster-level VLM judgments and then manually reviewed against the final rendered
posters; the reported percentages use the reviewed labels. The categories are
not mutually exclusive, and one poster may contribute to multiple categories.
These measurements characterize observable final-output failures rather than
internal checkpoint activity or contract-compliance status.

For the PPTX comparison in Figure~4 of the main paper, we apply fixed rules to
poster-level
VLM-as-Judge outputs for paired PosterGen and \ourmethod-PPT posters. Severe
text crowding requires an engagement score of at most 2 and an explanation
explicitly identifying dense or excessive text. A content gap requires a
content-completeness score of at most 3 and an explanation identifying missing
or underdeveloped content. An illegible figure requires a relevant score of at
most 3 and an explanation identifying a small or unreadable visual. A
reading-flow disruption is counted only when the explanation explicitly
identifies a broken order or narrative flow. Categories are not mutually
exclusive. All rule-flagged signals are manually reviewed against the final
rendered posters, and the reported percentages use the reviewed labels. These
measurements capture VLM-observable output signals and are not direct
evaluations of executable assertions in $\mathcal{E}$.

\subsection{PosterGen Transfer Set}
\label{sec:transfer}

We evaluate transfer on the 30-paper set provided by the PosterGen authors.
Because it lacks PaperQuiz questions, we generate 50 verbatim and 50
interpretive questions per paper once and freeze them before comparing methods.

GPT-4.1 generates the automatic posters, GPT-4o provides all VLM-as-Judge
scores, and GPT-4o, GPT-4o-mini, and o3 form the image-only PaperQuiz ensemble.
All methods share the frozen questions, prompts, scoring, and aggregation;
density augmentation uses the main-paper formula with the transfer benchmark's
separately computed GT-Poster reference median $w=1335.0$. The 100-paper primary
and cross-model experiments instead use their benchmark-specific median
$w=1335.5$.

\begin{table*}[t]
\centering
\scriptsize
\setlength{\tabcolsep}{1.45pt}
\resizebox{\textwidth}{!}{%
\begin{tabular}{@{}l*{15}{c}@{}}
\toprule
\multirow{3}{*}{Method}
& \multicolumn{9}{c}{VLM-as-Judge ($\uparrow$)}
& \multicolumn{6}{c}{PaperQuiz ($\uparrow$)} \\
\cmidrule(lr){2-10} \cmidrule(lr){11-16}
& \multicolumn{4}{c}{Aesthetic}
& \multicolumn{4}{c}{Information}
& \multirow{2}{*}{Overall}
& \multicolumn{3}{c}{Raw PaperQuiz Accuracy}
& \multicolumn{3}{c}{Density-Augmented PaperQuiz} \\
\cmidrule(lr){2-5} \cmidrule(lr){6-9}
\cmidrule(lr){11-13} \cmidrule(lr){14-16}
& Elem. & Layout & Engage. & Avg.
& Clarity & Content & Logic & Avg. &
& Verbatim & Interpretive & Overall
& Verbatim & Interpretive & Overall \\
\midrule
GT Poster
& 3.97 & 3.73 & 3.10 & 3.60 & 4.40 & 4.00 & 4.37 & 4.26 & 3.93
& 71.27 & 87.93 & 79.60 & 135.67 & 167.80 & 151.73 \\
\midrule
PosterAgent
& \textbf{4.03} & \underline{3.70} & \textbf{3.13} & \underline{3.62} & 3.90 & 4.00 & \underline{4.23} & 4.04 & 3.83
& 70.07 & 90.00 & 80.03 & 134.12 & 172.36 & 153.24 \\
PosterGen
& \textbf{4.03} & 3.60 & \underline{3.07} & 3.57 & 4.53 & 4.00 & \underline{4.23} & 4.26 & \underline{3.91}
& 65.60 & \textbf{92.27} & 78.93 & 130.44 & \textbf{183.45} & 156.94 \\
PPTAgent
& 2.07 & 3.40 & 2.53 & 2.67 & 4.33 & 3.07 & 3.67 & 3.69 & 3.18
& 24.80 & 46.87 & 35.83 & 49.60 & 93.73 & 71.67 \\
P2P
& 3.73 & 3.37 & \underline{3.07} & 3.39 & 4.50 & 4.03 & \underline{4.23} & 4.26 & 3.82
& 67.60 & 87.73 & 77.67 & 133.07 & 172.81 & 152.94 \\
PosterForest
& \underline{4.00} & 3.33 & 3.03 & 3.46 & \underline{4.57} & 4.00 & 4.13 & 4.23 & 3.84
& 70.67 & 89.20 & 79.93 & 135.56 & 171.00 & 153.28 \\
Any2Poster
& 2.00 & 2.33 & 2.33 & 2.22 & 4.03 & 3.43 & 3.53 & 3.67 & 2.94
& 53.53 & 81.40 & 67.47 & 107.07 & 162.80 & 134.93 \\
EfficientPosterGen
& 3.90 & 3.57 & 2.93 & 3.47 & 4.37 & 3.80 & 4.10 & 4.09 & 3.78
& 59.33 & 82.33 & 70.83 & 116.39 & 161.75 & 139.07 \\
PosterHarness
& 3.07 & 2.77 & 2.27 & 2.70 & 4.07 & \textbf{4.33} & 3.30 & 3.90 & 3.30
& 66.67 & 87.73 & 77.20 & 133.33 & 175.47 & 154.40 \\
\midrule
\ourmethod-PPT
& \underline{4.00} & \textbf{3.83} & \underline{3.07} & \textbf{3.63} & \textbf{4.87} & 4.00 & \textbf{4.30} & \textbf{4.39} & \textbf{4.01}
& \textbf{82.60} & 88.40 & \textbf{85.50}
& \textbf{164.23} & 175.88 & \textbf{170.06} \\
\ourmethod-HTML
& 3.90 & 3.23 & \textbf{3.13} & 3.42 & 4.53 & \underline{4.10} & \underline{4.23} & \underline{4.29} & 3.86
& \underline{71.00} & \underline{91.73} & \underline{81.37}
& \underline{137.33} & \underline{177.11} & \underline{157.22} \\
\bottomrule
\end{tabular}
}
\caption{Transfer results on the 30-paper PosterGen evaluation set. Automatic
posters use GPT-4.1 for generation; all VLM-as-Judge scores in this table are
produced by GPT-4o using the same six criteria as in the main experiment.
PaperQuiz uses the shared
three-reader image-only ensemble of GPT-4o, GPT-4o-mini, and o3; scores are
averaged across readers per poster and then across papers. We report Raw
PaperQuiz Accuracy and per-poster Density-Augmented PaperQuiz scores using the fixed reference
$w=1335.0$. The GT Poster is included only as a reference and is excluded from
automatic-method ranking. Bold and underlined values denote the best and
second-best automatic results; rounded ties share the same mark.}
\label{tab:transfer-full}
\end{table*}

\begin{table*}[t]
\centering
\small
\caption{Sensitivity analyses for pairwise preference. Holm $p$ adjusts both
tests; bootstrap CIs cross-cluster by annotator and paper.}
\label{tab:human-cluster-pair}
\setlength{\tabcolsep}{12pt}
\begin{tabular}{@{}l c c c@{}}
\toprule
Comparison & Holm $p$ & Bootstrap CI & LOO range \\
\midrule
PPT vs. PosterGen & .0015 & 62.5--82.4\% & 70.6--74.6\% \\
HTML vs. P2P & .278 & 46.3--63.7\% & 53.0--57.2\% \\
\bottomrule
\end{tabular}

\par\smallskip
\begin{minipage}{0.76\textwidth}
\small
Leave-one-annotator-out results do not attribute the PPTX preference to one
annotator. The HTML intervals include 50\%, so that result remains descriptive.
\end{minipage}
\end{table*}

\section{Evaluation and Human-Study Protocols}
\label{sec:supp-evaluation}
\subsection{Automatic Evaluation Protocol}
\label{sec:evaluation}
The primary protocol uses the 100-paper Paper2Poster benchmark; cross-model and
transfer exceptions are defined in Sections~\ref{sec:cross-model}
and~\ref{sec:transfer}. Each system produces one poster per paper without
independent candidate selection. HTML and PPTX outputs are rendered with
Playwright and LibreOffice, respectively, and all automatic metrics evaluate
only the final poster image.

\paragraph{Automatic metrics.}
For external VLM-as-Judge evaluation, all comparisons use GPT-4o with the
benchmark's six criteria and five-level anchors. PaperQuiz uses 50 frozen
verbatim and 50 frozen interpretive
questions per paper, an image-only prompt, common scoring, and the same ensemble
of GPT-4o, GPT-4o-mini, and o3. Scores are averaged across readers per paper and
then across the benchmark; missing, invalid, or unsupported answers are
incorrect. For the 100-paper primary and cross-model experiments, density
augmentation follows the main-paper formula with the fixed GT-Poster median
$w=1335.5$; the 30-paper transfer exception is defined in
Section~\ref{sec:transfer}.

\subsection{Human Evaluation Protocol}
\label{sec:human}
Fifteen trained annotators with AI research backgrounds completed the seven-way
ranking task, and 13 also completed the paired comparisons. Method identities
were concealed; poster order and pairwise left--right placement were randomized.
The seven-source subset was fixed when the study was run; baselines added later
to the main comparison were omitted because their public implementations were
unavailable or not reproducible at that time.

\paragraph{Seven-way ranking.}
For mean-rank calculation in the seven-way task, unselected systems share the
average of the remaining ranks; when three systems are selected, each of the
other four receives rank 5.5. Across 1{,}175 trials, annotators provide 1{,}161
top-1, 1{,}157 top-2, and 1{,}139 top-3 selections, with the differences arising
from permitted abstentions. Restricting the sensitivity analysis to the 11
annotators who completed all 100 papers preserves the direction of every paired
mean-rank difference.

\paragraph{Within-pipeline pairs.}
The study contains 1{,}296 PPTX and 1{,}292 HTML comparisons. Because annotators
and papers are repeatedly observed, primary 95\% confidence intervals use
two-way cluster-robust standard errors~\cite{cameron2011robust}. A
20{,}000-replicate crossed-cluster bootstrap provides sensitivity intervals,
and Holm correction is applied across the two tests.

\clearpage
\section{Extended Qualitative Results}
\label{sec:supp-qualitative}

\begin{figure*}[p]
\captionsetup{skip=2pt}

\noindent\begin{minipage}{\textwidth}
\centering
\includegraphics[width=0.90\linewidth]{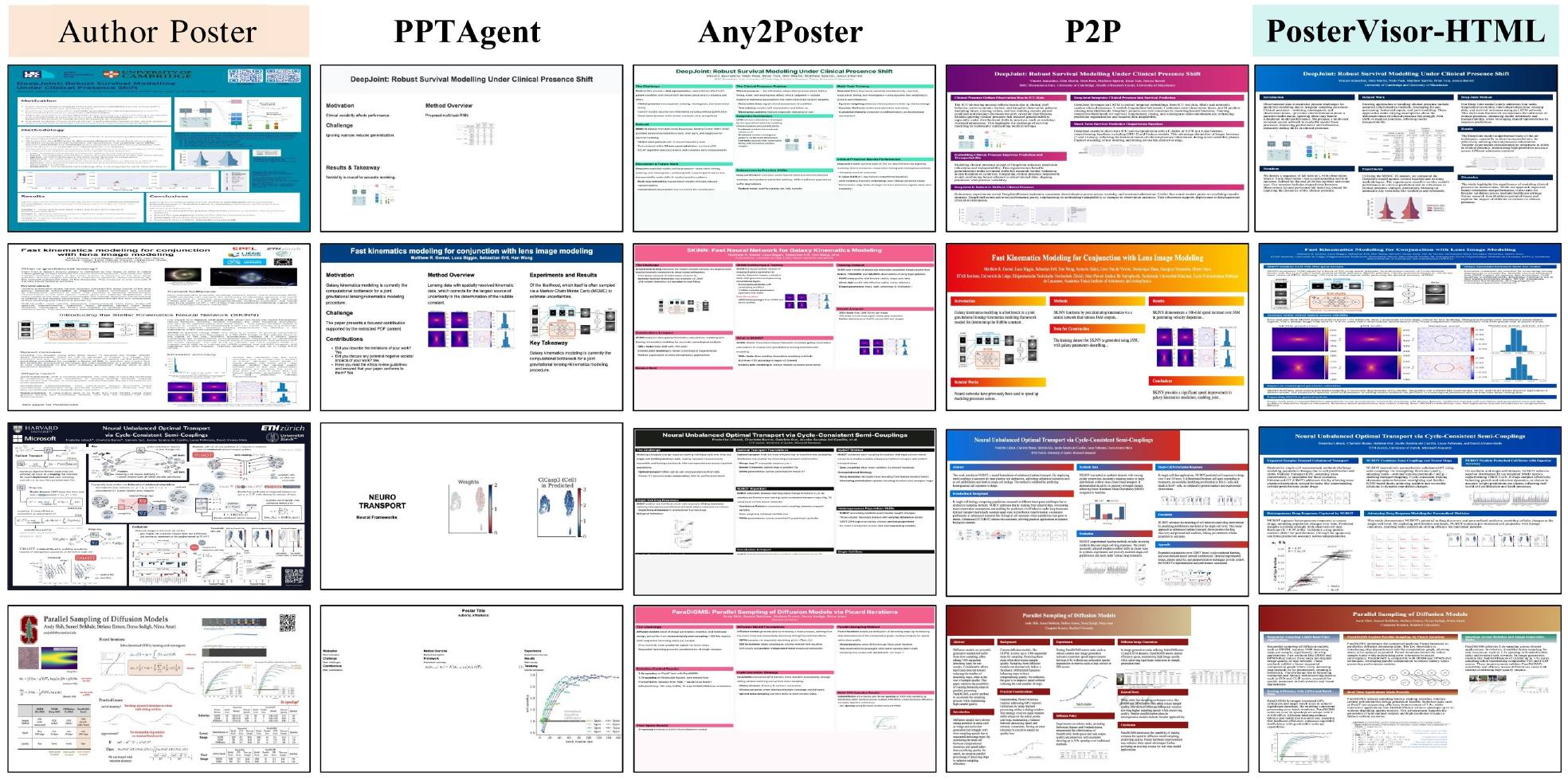}
\captionof{figure}{Representative HTML-oriented outputs on four papers. Columns
show Author Poster, PPTAgent, Any2Poster, P2P, and \ourmethod-HTML.}
\label{fig:supp-case-1}
\end{minipage}

\vspace{0.05em}

\noindent\begin{minipage}{\textwidth}
\centering
\includegraphics[width=0.90\linewidth]{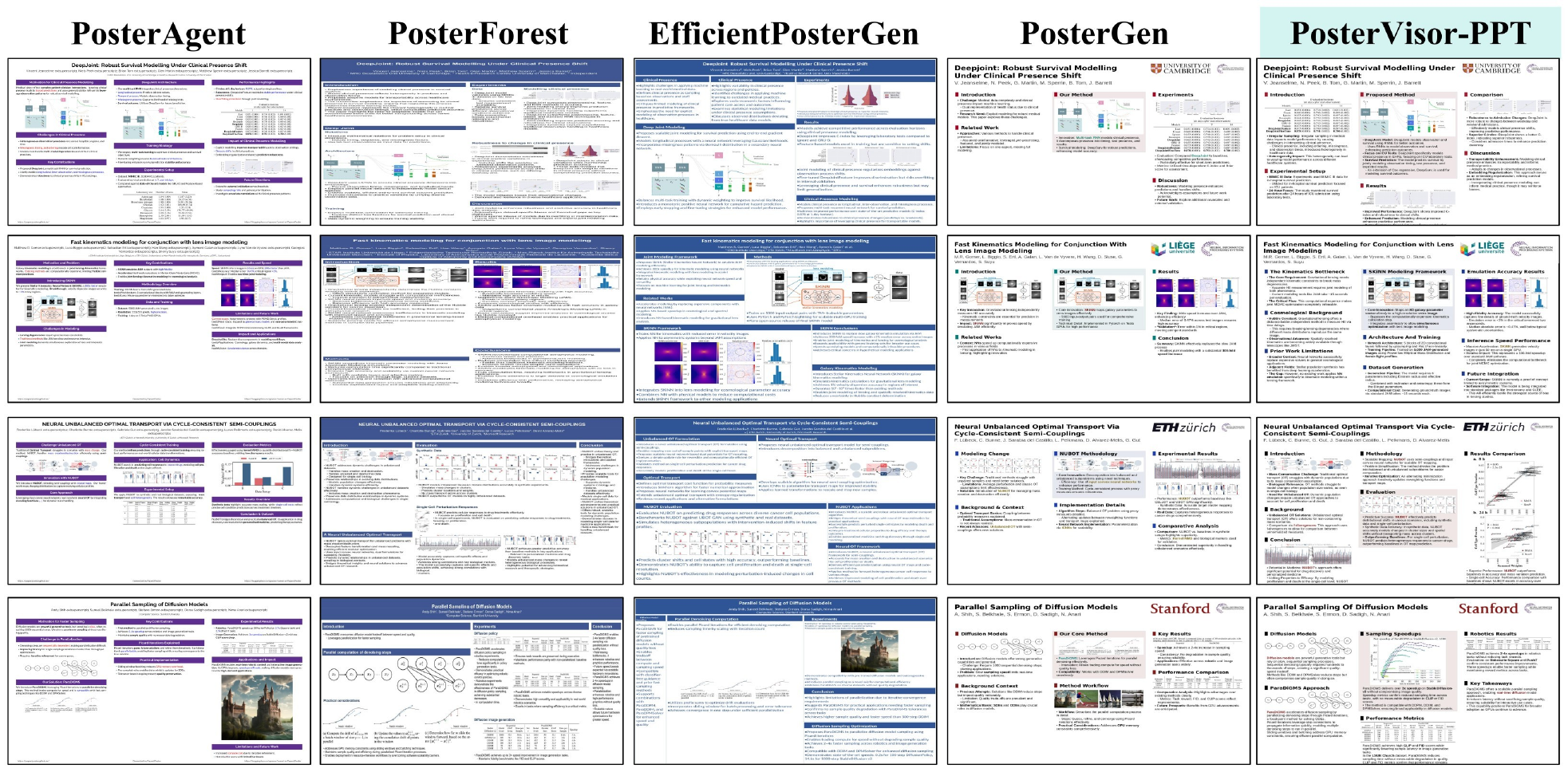}
\captionof{figure}{Representative PPTX-oriented outputs on the same four papers.
Columns show PosterAgent, PosterForest, EfficientPosterGen, PosterGen, and
\ourmethod-PPT.}
\label{fig:supp-case-2}
\end{minipage}

\vspace{0.05em}

\noindent\begin{minipage}{\textwidth}
\centering
\includegraphics[width=0.63\linewidth]{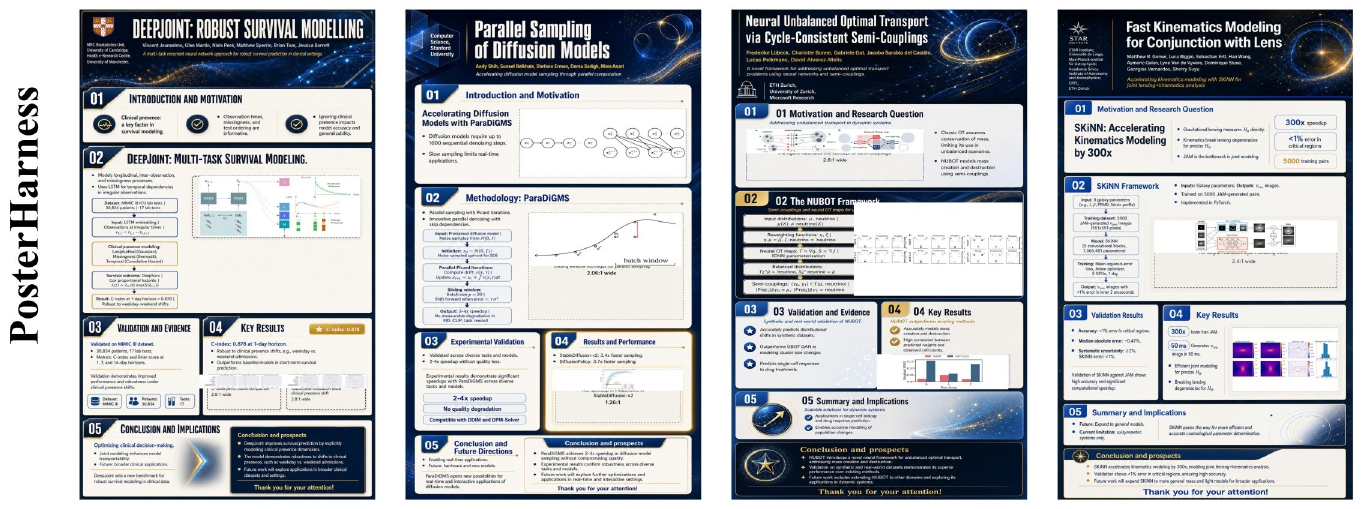}
\captionof{figure}{Representative PosterHarness outputs on the same four papers.}
\label{fig:supp-case-3}
\end{minipage}

\end{figure*}

Figures~\ref{fig:supp-case-1} and~\ref{fig:supp-case-2} compare HTML- and
PPTX-oriented methods on the same four papers. They support direct inspection of
density, evidence use, readability, and spatial organization, but not aggregate
performance estimates or component-level causal attribution. Light-blue headers
identify \ourmethod outputs. Figure~\ref{fig:supp-case-3} additionally shows
PosterHarness outputs on the same paper set.

\newpage
\raggedbottom
\begingroup
\interlinepenalty=10000
\bibliography{aaai2027}
\endgroup

